\documentclass[sigconf]{aamas} 

\usepackage{balance} 
\usepackage{tabularx}
\usepackage{multirow}
\usepackage{amsmath}
\usepackage{graphicx}
\usepackage{caption}
\usepackage{subcaption}
\usepackage{makecell}
\usepackage{array}
\usepackage{longtable}

\newcolumntype{P}[1]{>{\arraybackslash}p{#1}}

\setcopyright{ifaamas}
\acmConference[AAMAS '24]{Proc.\@ of the 23rd International Conference
on Autonomous Agents and Multiagent Systems (AAMAS 2024)}{May 6 -- 10, 2024}
{Auckland, New Zealand}{N.~Alechina, V.~Dignum, M.~Dastani, J.S.~Sichman (eds.)}
\copyrightyear{2024}
\acmYear{2024}
\acmDOI{}
\acmPrice{}
\acmISBN{}

\acmSubmissionID{384}

\title{Learning and Sustaining Shared Normative Systems \\ via Bayesian Rule Induction in Markov Games}

\author{Ninell Oldenburg}
\affiliation{
  \institution{University of Copenhagen}
  \city{}
  \country{}
}
\email{niol@hum.ku.dk}

\author{Tan Zhi-Xuan}
\affiliation{
  \institution{Massachusetts Institute of Technology}
  \city{}
  \country{}
}
\email{xuan@mit.edu}

\begin{abstract}
A universal feature of human societies is the adoption of systems of rules and norms in the service of cooperative ends. How can we build learning agents that do the same, so that they may flexibly cooperate with the human institutions they are embedded in? We hypothesize that agents can achieve this by assuming there exists a shared set of norms that most others comply with while pursuing their individual desires, even if they do not know the exact content of those norms. By assuming shared norms, a newly introduced agent can infer the norms of an existing population from observations of compliance and violation. Furthermore, groups of agents can converge to a shared set of norms, even if they initially diverge in their beliefs about what the norms are. This in turn enables the stability of the normative system: since agents can bootstrap common knowledge of the norms, this leads the norms to be widely adhered to, enabling new entrants to rapidly learn those norms. We formalize this framework in the context of Markov games and demonstrate its operation in a multi-agent environment via approximately Bayesian rule induction of obligative and prohibitive norms. Using our approach, agents are able to rapidly learn and sustain a variety of cooperative institutions, including resource management norms and compensation for pro-social labor, promoting collective welfare while still allowing agents to act in their own interests.
\end{abstract}


\newcommand{\BibTeX}{\rm B\kern-.05em{\sc i\kern-.025em b}\kern-.08em\TeX}

\usepackage{balance}

\makeatletter
\gdef\@copyrightpermission{
	\begin{minipage}{0.3\columnwidth}
		\href{https://creativecommons.org/licenses/by/4.0/}{\includegraphics[width=0.90\textwidth]{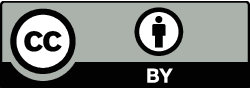}}
	\end{minipage}\hfill
	\begin{minipage}{0.7\columnwidth}
		\href{https://creativecommons.org/licenses/by/4.0/}{This work is licensed under a Creative Commons Attribution International 4.0 License.}
	\end{minipage}
	\vspace{5pt}
}
\makeatother

\begin{document}

\setlength{\abovedisplayskip}{1pt}
\setlength{\belowdisplayskip}{1pt}


\pagestyle{fancy}
\fancyhead{}


\maketitle 


\section{Introduction}\label{sec:introduction}

For autonomous agents to integrate with society, they will have to learn to comprehend and comply with the unspoken rules and shared expectations that pervade human interaction: social norms \cite{bicchieri_grammar_2005}. Such norms function as a way of enforcing common standards and promoting cooperative behavior \cite{gintis_social_2010,binmore1994economist, nyborg_social_2016,dagostino_contemporary_2021}, and thus serve as promising targets for aligning autonomous systems with societal interests even while they primarily serve the goals of their users \cite{arnold2017value,nay_law_2022,zhi-xuan_what_2022}.
How then can we build agents that learn and comply with norms? In answering this question, we need to contend with a paradoxical aspect of social normativity: Even though social norms function as \emph{shared} constraints on behavior \cite{tuomela_importance_1995,dagostino_contemporary_2021}, they are primarily learned and sustained in a \emph{decentralized} and \emph{emergent} manner \cite{axelrod1981evolution,ostrom1990governing,morris-martin_norm_2019,hawkins_emergence_2019}. As such, a useful model of norm learning should capture its shared-yet-decentralized nature, enabling agents to rapidly coordinate in new contexts without central control \cite{agapiou_melting_2023,hadfield_what_2012}.

In this paper, we develop a Bayesian account of social norm learning which directly addresses this paradox. Drawing upon frameworks for joint intentionality \cite{tomasello2007shared,tang_bootstrapping_2011,wu_too_2021} and shared agency \cite{tuomela_importance_1995,bratman_shared_2013} in human cooperation, we model agents that assume \emph{shared normativity} when learning to interact with others. In particular, each agent assumes that all other agents may be complying with a shared set of norms even while pursuing their own interests, allowing them to infer the existence of a norm from apparent violations of self-interest, and to aggregate such observations across a population of agents, updating their beliefs about which norms best explain collective behavior. Having inferred these norms, an agent can then comply with them, either because of an intrinsic motivation \cite{li2021young}, or because it is strategically valuable \cite{morris2018common}. Under sufficiently high levels of compliance, shared normativity is thus \emph{self-sustaining}: Once enough agents infer a shared set of norms, they conform to them, thereby creating public knowledge that the norms are practiced, and enabling new entrants to rapidly learn the norms.

\begin{figure*}[t]
\centering
    \begin{subfigure}[b]{0.42\textwidth}
        \includegraphics[width=\textwidth]{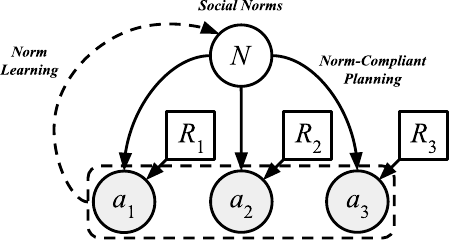}
        \subcaption{\textmd{Norm-Augmented Markov Game}}
        \label{subfig:graphical_model}
    \end{subfigure}
    \hspace{0.06\textwidth}
    \begin{subfigure}[b]{0.38\textwidth}
        \includegraphics[width=\textwidth]{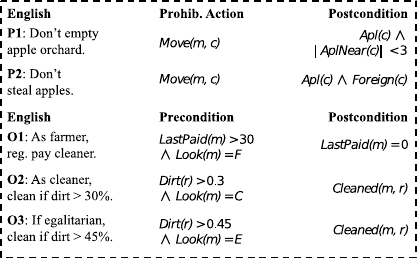}
        \subcaption{\textmd{Social Norms}}
        \label{subfig:social_norms}
    \end{subfigure}
    
    \begin{subfigure}[b]{0.4\textwidth}
        \includegraphics[width=\linewidth]{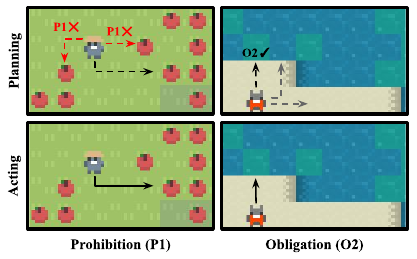}
        \subcaption{\textmd{Norm-Compliant Planning}}
        \label{subfig:norm_compliant_planning}
    \end{subfigure}
    \hspace{0.08\textwidth}
    \begin{subfigure}[b]{0.4\textwidth}
        \includegraphics[width=\linewidth]{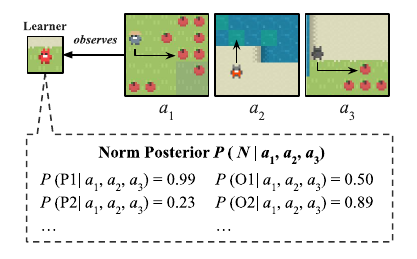}
        \subcaption{\textmd{Norm Learning}}
        \label{subfig:bayesian_norm_learning}
    \end{subfigure}
    \caption{Overview of our framework,
    \textmd{summarized by \textbf{(a)} the (simplified) graphical model for a norm-augmented Markov game. Agents learn to comply with rule-based social norms \textbf{(b)}, which are factored into prohibitions (e.g. P1, P2) and obligations (e.g. O1, O2, O3). Agents comply with norms while pursuing their interests via model-based planning \textbf{(c)}, which switches between a reward-oriented mode that obeys prohibitions while collecting rewards (left), and an obligation-oriented mode that plans to satisfy a postcondition (right). Agents learn these norms via \textbf{(d)} approximate Bayesian inference, updating the probabilities of each potential norm based on observations of others' actions.}}
    \label{fig:components}
\end{figure*}

We formalize this account in the context of multi-agent sequential decision-making, introducing \emph{norm-augmented Markov games} as a framework for studying norm-guided learning and planning. In contrast to similar frameworks based on (model-free) reinforcement learning (RL) \cite{vinitsky_learning_2023,lerer_learning_2019,jaques2019social}, our approach foregrounds the \emph{epistemic} and \emph{cognitive} aspects of social norms, framing norm learning as an (approximately) rational Bayesian process of learning structured rules \cite{goodman_rational_2008,nichols2016rational,savarimuthu_identifying_2013,craneeld_bayesian_2016,zhi-xuan_thats_2019}, rather than a reactive process of adapting one's habitual actions \cite{koster_model-free_2020,ayars2016can}. As such, our account captures the speed and flexibility of norm learning and coordination in human children and adults \cite{rakoczy_sources_2008,rakoczy_young_2009,schmidt_young_2011}, while providing a normative standard against which other approaches can be measured. We instantiate and evaluate this framework in DeepMind's Melting Pot simulator \cite{leibo_scalable_2021, agapiou_melting_2023}, demonstrating that approximately Bayesian norm learning is indeed feasible in long-horizon multi-agent contexts while requiring orders of magnitude less experience than model-free norm learning from sanctions \cite{koster_model-free_2020,vinitsky_learning_2023}. We also find that shared normativity enables both the maintenance and convergence of normative systems: Common knowledge of the norms is maintained despite the ``deaths'' of experienced agents and the ``births'' of normative novices, while groups of agents can bootstrap themselves towards a stable set of norms by learning from each others' actions and experimentally complying with tentative norms.

Collectively, these findings highlight the importance of integrating game-theoretic accounts of norms and conventions \cite{axelrod1981evolution,binmore1994economist,gintis_social_2010} with the conceptual richness and flexibility of human learning \cite{goodman_rational_2008,goodman2014concepts,piantadosi2016four}. By uniting the structured representations of normative concepts common in normative systems research \cite{boella_introduction_2006,boella_architecture_2006,camilleri_contracts_2017} with a Bayesian decision-theoretic approach to multi-agent coordination \cite{aumann_correlated_1987,kalai1995subjective}, our framework models the dynamism of human normative cognition that is omitted from model-free RL approaches, while paving the way for normatively-competent autonomous systems that combine model-free and model-based learning \cite{cushman2013action,levine_resource-rational_2023}.

\section{Learning and Sustaining Normative Systems in Markov Games}\label{sec:modelingnorms}
How can we design agents that learn and maintain systems of social norms? As motivated above, we first introduce \emph{norm-augmented Markov games} (NMGs) as an extension of Markov games \cite{littman_markov_1994, gmytrasiewicz_framework_2005, perolat_multi-agent_2017, liu_sample-efficient_2022, christoffersen_get_2022}, providing a formal setting for norm learning and coordination in sequential decision-making (Figure \ref{subfig:graphical_model}). As an instance of such a game, we adapt the Melting Pot simulator \cite{leibo_scalable_2021, agapiou_melting_2023} to create a multi-agent environment that affords a range of cooperative norms. We then introduce a rule-based representation of social norms (Figure \ref{subfig:social_norms}), which take the form of obligations or prohibitions, and show how agents can comply with these norms through model-based planning (Figure \ref{subfig:norm_compliant_planning}). Finally, we show how agents can perform approximately Bayesian \emph{rule induction} \cite{muggleton_bayesian_1994, goodman_rational_2008}, inferring norms from observations of violation and compliance with potential obligations and prohibitions (Figure \ref{subfig:bayesian_norm_learning}).

\subsection{Norm-Augmented Markov Games}\label{sec:norm_augmentedMG}

In Markov games \cite{littman_markov_1994, gmytrasiewicz_framework_2005, perolat_multi-agent_2017, liu_sample-efficient_2022}, a set of $M$ agents take actions $a_i \in \mathcal{A}_i$ in a series of environment states $s \in \mathcal{S}$ over time, where $i \in \mathcal{I} := \{1, ..., M\}$ is each agent's index. Actions change the environment state via a transition function $T: \mathcal{S} \times \mathcal{A}_1 \times ... \times \mathcal{A}_M \to \Delta(\mathcal{S})$ that maps the current state $s$ and joint actions $(a_1, ..., a_M)$ to a distribution over successor states $s'$. Each agent also has a reward function $R_i: \mathcal{S} \times \mathcal{A}_i \times \mathcal{S} \to \mathbb{R}$ that maps state transitions $(s, a, s')$ to scalar rewards $r \in \mathbb{R}$. We also assume a shared discount factor $\gamma$. In our context, an agent's reward function can be understood as a representation of their individual desires. Markov games begin with an initial state $s_1$, and end after a horizon of $H$ steps.

We extend Markov games with the addition of \emph{social norms}, formalized as functions $\nu: \mathcal{H} \to \{0, 1\}$ where $\mathcal{H}_i := \bigcup_{t=1}^H ((\mathcal{S} \times \mathcal{A}_i)^t \times \mathcal{S})$ is the set of state-action histories for agent $i$, and $\mathcal{H} := \bigcup_{i \in \mathcal{I}} \mathcal{H}_i$ is the set of histories across all agents. In words, a social norm $\nu$ \emph{classifies} whether the behavioral history of an agent $i$ complies with the norm \cite{hadfield_what_2012}, with 0 indicating compliance and 1 indicating violation. We call a set of norms $N$ a \emph{normative system}.

Let $\mathcal{N}$ be a set of functions defining the space of possible social norms, and $\epsilon \in [0, 1]$ be the frequency of a norm violation. We now define norm-augmented Markov games as a (subjective) Bayesian extension of a Markov game, endowing each agent $i$ with a subjective prior $P^0_i(N)$ over which sets of norms $N \subseteq \mathcal{N}$ will be complied with by \emph{all agents} with frequency at least $1 - \epsilon$:
\begin{equation}
    P^0_i(\nu) = P_i\left(\frac{1}{HM}\sum_{t,j=1}^{H, M} (1 - \nu(h^t_j)) \geq 1 - \epsilon\right), \quad
    P^0_i(N) = \prod_{\nu \in N} P^0_i(\nu)
    \label{eq:norm-prior}     
\end{equation}
where $h^t_j$ is the state-action history for agent $j$ at step $t$, treated as a random variable.
Each agent also has conditional prior $P_i(h_t|N)$ over agent histories $h_t$ given norms $N$, and this induces a sequence of posteriors over norms $P_i^t(N) := P_i(N | h_t) \propto P^0_i(N) P_i(h_t | N)$, defining agent $i$'s belief in norms $N$ at each $t$ after observing $h_t$. We also endow each agent $i$ with a norm violation cost function $C_i: \mathcal{N} \to \mathbb{R}$, which defines how intrinsically motivated an agent is to comply with norms that they believe to be true. This induces a \emph{norm-augmented reward function} $R'_i: \mathcal{H} \to \mathbb{R}$. Given a history $h = (h', s, a_i, s')$, $R'_i$ decomposes into a base reward function $R_i$ and the expected cost of norm violation under the posterior $P_i(N | h_t)$:

\begin{equation}
    R'_i(h', s, a_i, s') := R_i(s, a_i, s') - \sum_{N \subseteq \mathcal{N}} P_i(N | h) \sum_{\nu \in N} C_i(\nu) \nu(h)
    \label{eq:norm-augmented-reward}
\end{equation}
\vspace{4pt}

By adding priors over norms to a Markov game, each agent $i$ effectively factors their beliefs about other agent's policies $\pi_{-i}$ into two parts: Beliefs about whether the policies will comply with a \emph{shared} set $N$ of relatively simple yes/no rules, and beliefs about non-normative features of those policies. This is illustrated in Figure \ref{subfig:graphical_model}, depicting how agents' actions $a_i$ are influenced by both shared norms $N$ and individual reward functions $R_i$. An agent's prior over norms thus serves as a \emph{correlating device} over their beliefs about how other agents behave \cite{aumann_correlated_1987,gintis_social_2010}. If agents then act according to these correlated beliefs (e.g. by complying with the norms), then their \emph{actual} policies will also be correlated \cite{kalai1995subjective}, helping agents achieve coordination. The existence of norm violation costs $C_i$ further aids this coordination, promoting compliance with a normative system $N$ once it is sufficiently believed in.

When is a normative system $N$ stable? In general, a set of beliefs about norms $\{P_i^t(N)\}_{i=1}^M$ at time $t$ is consistent with a wide range of correlated strategies $\pi_{1:M}$, some of which may form (subjective) correlated equilibria: joint policies which no agent has an incentive to deviate from \cite{aumann_correlated_1987,liu_sample-efficient_2022}. Out of these equilibria, consider those where the \emph{expected frequency} of each agent $i$ complying with norms $N \sim P_i^t(N)$ over all future steps is at least $1-\epsilon$, where the expectation is taken over their own belief distribution $P_i^t(N)$ over the norms. Since these are policies where agents frequently comply with the norms they (subjectively) believe might be true, we call them \emph{subjective normative equilibria}. If all agents have converged to a single distribution $P^t(N)$ over normative systems (or $\delta$-close to it), this forms an \emph{(objective) normative equilibrium} (respectively, a normative $\delta$-equilibrium). If $P^t(N)$ \emph{concentrates} on a specific $N$, then $N$ is \emph{stable} with respect to that normative equilibrium. Note that if $P^t(N)$ is concentrated enough on some $N$, and agents are (norm-augmented) reward-maximizers, we can guarantee that $N$ is stable by setting violation costs for $\nu \in N$ to be sufficiently high.

As an example setting for norm-augmented Markov games, we adapt and combine several existing games from the Melting Pot suite of environments \cite{leibo_scalable_2021, agapiou_melting_2023}, including elements of Commons Harvest \cite{janssen_lab_2010, perolat_multi-agent_2017}, Clean Up \cite{hughes_inequity_2018}, and Territory \cite{leibo_scalable_2021}. The resulting environment is shown in Figure \ref{fig:components}. In the environment, agents can collect and eat apples to gain reward but otherwise incur a cost for actions. Apples regrow at a density-dependent rate (more surrounding apples lead to faster regrowth), but the rate decreases as the nearby river gets polluted over time. To prevent pollution, agents have to use clean-up action on the river. In addition, several regions are visually marked as each agent's territory (but do not otherwise have functional effects), and agents have one of three visually distinct appearances that demarcate \textit{roles}. By combining these elements, we allowed for a range of social norms, including resource management norms, property norms, role-based division of labor, and compensation for pro-social labor. In our experiments, we gave experienced agents high prior beliefs in some norm set $N$, and learning agents a low prior belief in all (non-empty) norm sets. We present details of the environment's dynamics in the Appendix.

\subsection{Representing Social Norms}\label{sec:rules}

NMGs are a general setting for multi-agent norm learning, but they leave unspecified the set $\mathcal{N}$ of possible norms. Drawing upon deontic and temporal logic representations of norms \cite{alchourron_logic_1969,craneeld_bayesian_2016,kasenberg_inverse_2018}, we consider learning both (maintenance-based) \emph{prohibitions} $\mathcal{N}_P$, which forbid the effects of certain actions at all times, and (achievement-based) \emph{obligations} $\mathcal{N}_O$, which require pursuing \textit{temporally extended} goals, giving our full set of norms $\mathcal{N} = \mathcal{N}_P \cup \mathcal{N}_O$. Several examples are shown in Figure~\ref{subfig:social_norms}. Note that the distinction between prohibitions and obligations is not fundamental: Prohibitions can be represented as obligations via negating the effects and vice versa. Nonetheless, we found it more natural to represent maintenance norms in their prohibitive form (e.g. ``Don't empty the apple orchard.'') and achievement norms in their obligative form (e.g. ``Pay the cleaner in the next 30 steps.'').

Following prior work on norm and rule learning \cite{zhi-xuan_thats_2019,goodman_rational_2008}, we express norms as rules in a first-order language $\mathcal{L}$, which includes predicates and functions defined over objects in a state $s$. Given a first-order expression $\phi$, we denote its valuation in $s$ as $s[\phi]$. Using this notation, we represent a prohibition norm $\nu_P \in \mathcal{N}_P$ as a tuple $\langle \textsf{Prohib}, \textsf{Post} \rangle$ where $\textsf{Prohib}$ is a function of objects in $s$ that returns a set of potentially prohibited actions, and the postcondition $\textsf{Post}$ holds in a state $s$ whenever the prohibited effect occurs. In other words, if an action $a_i \in s[\textsf{Prohib}]$ leads to state $s'$ such that $s'[\textsf{Post}]$ is true, this means that the prohibition has been violated, and hence that $\nu_{P}(h, s, a, s') = 1$. As an example, an agent \textsf{m} in our environment is prohibited from moving to a cell \textsf{c} ($\textsf{Prohib}=\textsf{Move(m, c)}$) if that cell contains apples and is in foreign territory ($\textsf{Post} = \textsf{Apl(c)} \land \textsf{Foreign(c)}$) as depicted in Figure~\ref{subfig:social_norms}.

Next, we define obligations $\nu_O \in \mathcal{N}_O$ as tuples $\langle \textsf{Pre}, \textsf{Post}, \tau \rangle$, where $\textsf{Pre}$ is a precondition formula that holds true when the obligation is triggered, and $\textsf{Post}$ is a postcondition formula that holds true once the obligation has been discharged, with $\tau$ as the duration during which the obligation must be satisfied. Given a history $h=(s_{t-k}, a_{-i,t-k}, ..., s_t, a_{n,t})$ up to time step $t$, this means that $\nu_O(h) = 1$ (the obligation is violated) if there exists some $k \leq \tau$ such that $s_{t-k}[\textsf{Pre}]$ is true but $s_{t-k+j}[\textsf{Post}]$ is false for all $k < j \leq \tau$. As another example, when agents are in the cleaner role ($\textsf{Look(m) = C}$) and find the river more than 30\% polluted ($\textsf{Dirt(r) > 0.3}$) they are obliged to have cleaned the river ($\textsf{Cleaned(m, r)}$) within the next 20 steps ($\tau = \textsf{20}$). Therefore, ($\textsf{Pre}=(\textsf{Dirt(r) > 0.3} \land \textsf{Look(m) = C})$, $\textsf{Post} = \textsf{Cleaned(m, r)}$, $\tau = \textsf{20}$).

\subsection{Norm-Compliant Planning}\label{sec:planning}

Having introduced how norms classify behavior, we now describe how agents can plan to comply with a set of norms $N = N_P \cup N_O$ where $N_P \subseteq \mathcal{N}_P$ and $N_O \subseteq \mathcal{N}_O$. Absent strategic considerations (which norms themselves are meant to solve), each agent's goal is to maximize the sum of rewards given by the norm-augmented reward function (Equation \ref{eq:norm-augmented-reward}). We first assume that the agent is certain that a set of norms $N$ holds, then discuss how agents can plan under uncertainty about which norms are true.

Even when an agent is certain about $N$, complying with all of them can be challenging due to temporally extended obligations. As such, we design our agents to use two different planning modes, namely \emph{reward-oriented planning} and \emph{obligation-oriented planning}. Each of these corresponds to a different sub-problem that is a \emph{proxy} of the true game. Agents switch from the former mode to the latter mode whenever an obligation precondition is triggered.

In reward-oriented planning, agents maximize Equation \ref{eq:norm-augmented-reward} except with the cost of obligation norms ignored. Since prohibition norms are not temporally extended, this gives us a proxy reward $R^{N_P}_i(s, a_i, s')$ that only depends on the current state transition. We maximize this reward through model-based planning, assuming that each agent $i$ has a reasonably accurate model $T_i(s' | s, a_i)$ of the environment and other agents. In particular, we use real-time dynamic programming (RTDP) \cite{barto_learning_1995} to compute estimates of the state and action value functions:
\vspace{6pt}
\begin{align}
    Q^{N_P}_{i}(s, a_i) &=  \sum_{s' \in \mathcal{S}} T_i(s'| s, a_i) \left[ R^{N_P}_{i}(s, a_i, s') + \gamma V^{N_P}_{i}(s') \right] \label{eq:q-value-prohib} \\ 
    V^{N_P}_{i}(s) &= \max_{a_i \in \mathcal{A}_i} Q^{N_P}_{i}(s, a_i) 
    \label{eq:value-prohib}
\end{align}

While reward-oriented planning accounts for prohibitions, it ignores the potential future costs that might arise due to obligations becoming active. Instead, our agents reactively switch to obligation-oriented planning once the precondition $\textsf{Pre}$ of some obligation $\nu_O = \langle \textsf{Pre}, \textsf{Post}, \tau \rangle$ is met and then plan to achieve $\textsf{Post}$ as a goal. This creates a (stochastic) shortest-path problem, where any state $s$ with $s[\textsf{Post}]=1$ is treated as a \emph{terminal} state, with a modified reward function that accounts only for action costs $C_{a_i}$, prohibition costs $C_i(\nu_P)$ and delivers a reward of 1 when $\textsf{Post}$ is satisfied:
\vspace{6pt}
\begin{equation}
    R^{\nu_O \cup N_P}_i(s, a_i, s') = C_{a_i} +  \sum_{\nu_P \in N_P} C_i(\nu_P) \nu_P(s, a, s') + s'[\textsf{Post}]
    \label{eq:value-oblig}
\end{equation}
We similarly maximize this reward via model-based planning, giving us value estimates $Q^{\nu_O \cup N_P}_{i}$ and $V^{\nu_O \cup N_P}_{i}$ for each obligation $\nu_O$ that becomes active. Agents thus act by selecting actions that maximize either $Q^{N_P}_{i}$ or $Q^{\nu_O \cup N_P}_{i}$ depending on whether an obligation $\nu_O$ is active, performing several iterations of RTDP to update the value functions before acting. Obligations are satisfied in the order that their preconditions are triggered, and each agent maintains a queue of such obligations to satisfy.

Thus far, we have described how planning occurs when agents are \emph{certain} about a set of norms $N$. In general, however, agents only have a \emph{belief} $P_i^t(N)$ over norms. To handle this uncertainty, we consider two simple approaches, \emph{thresholding} and \emph{sampling}, which avoid the difficulty of taking expectations over all possible norm sets $N \subseteq \mathcal{N}$. When thresholding, agents comply with a norm $\nu$ whenever its probability is higher than a threshold $P_i^t(\nu \in N) \geq \theta$. When sampling, agents sample of set of norms $N' \sim P_i^t(N)$ to comply with for a certain number of steps $K$, before resampling. This is also known as probability matching \cite{gaissmaier2008smart} or Thompson sampling \cite{russo_tutorial_2018}. Each approach has properties that may be useful. Thresholding is low variance and may be appropriate when norm violation is sufficiently costly. In contrast, sampling can promote exploration, especially under high uncertainty about which norms are true \cite{eysenbach_if_2019}.

\subsection{Bayesian Norm Learning}\label{sec:learning}

As noted in Section \ref{sec:norm_augmentedMG}, a posterior $P_i^t(N)$ over norms is defined with respect to a conditional distribution $P_i(h_t |N)$ over how other agents will act, which we left unspecified. To specify this distribution, we make two assumptions: (i) Agents model each other as norm-compliant planners, and (ii) they assume that other agents act \emph{as if} they know the norms $N$ with certainty.
Since agents have full observability and can observe the actions $a_{-i}$ that all other agents take at state $s$ after history $h$. This induces the following posterior:
\begin{equation}
\label{eq:bayes_own}
P_i^{t+1}(N|h, s, a_{-i}) \propto \frac{\pi_{-i}(a_{-i}|h,s,N)\ P_i^t(N)}{P_i^t(a_{-i}|h,s)}
\end{equation}
The likelihood term $\pi_{-i}(a_{-i}|h,s,N)$ is the policy over actions that results from all other agents following the procedure in Section \ref{sec:planning}, conditional on all of them knowing that $N$ is true. This means that actions that comply with a norm $\nu \in N$ are more likely to occur, and violations are less likely. Note, however, that this may not be an accurate model in general. After all, other agents might \emph{also} be uncertain about the norms, and act based on their \emph{beliefs} about which norms are true. Nonetheless, the assumption of shared normativity greatly reduces the complexity of norm learning: Similar to how assumptions of shared agency make decentralized cooperation easier \cite{tang_bootstrapping_2011, ho_feature-based_2016, stacy_imagined_2022}, our agents avoid having to infer everyone else's beliefs about the norms, and directly infer the norms themselves.

Even with this assumption, inferring the exact posterior over $N$ --- the \emph{rule induction} problem \cite{muggleton_bayesian_1994,goodman_rational_2008} --- is still highly intractable, since it requires enumerating over all subsets $N' \subseteq \mathcal{N}$ and updating their probabilities. Instead, we compute a \emph{mean-field approximation} \cite{tanaka_theory_1998} of the posterior, assuming that it factorizes into approximate local posteriors $\tilde P^t(\nu|h, s, a_{-i})$ for each potential norm $\nu \in \mathcal{N}$:
\begin{equation}
\label{eq:mean-field}
P_i^{t+1}(N|h, s, a_{-i}) \propto \prod_{\nu \in \mathcal{N}} \tilde P^{t+1}(\nu|h, s, a_{-i})
\end{equation}

Making this approximation allows us to update beliefs about each norm $\nu$ \emph{independently} via Bayes rule:
\begin{equation}
\label{eq:local-posterior}
\tilde P^{t+1}(\nu|h, s, a_{-i}) \propto \pi_{-i}(a_{-i}|h,s,\nu) \tilde P_i^t(\nu)
\end{equation}
where $\pi_{-i}(a_{-i}|h,s,\nu)$ approximates the probability that the other agents take actions $a_{-i,t}$ under the assumption that $\nu$ is the \emph{only} norm that is true. We compute this probability by using the $Q$-values described in Section \ref{sec:planning}, assuming that each agent $j$ selects its action $a_{j}$ according to a Boltzmann distribution:
\begin{equation}
    \pi_{-i}(a_{j}|h,s,\nu) \propto \exp ( Q^{\nu}(s, a_j) )
\end{equation}
By using the $Q$-values derived by planning, we automatically take into account the fact that other agents are unlikely to violate norms since such actions will have lower $Q$-values.

\section{Experiments}\label{sec:findings}
Having developed our model of norm-guided learning and planning, we investigate whether it reproduces the aspects of human normativity that our theory suggests. To do, we conducted experiment studying four key phenomena: (1) \emph{Passive Norm Learning}: We explore the speed and effectiveness of learning norms via (approximately) Bayesian updating from passive observations \cite{craneeld_bayesian_2016}, assuming that norms are shared by other agents. (2) \emph{Norm-Enabled Social Outcomes}: We investigate whether our model of norm learning can foster collectively beneficial coordination \cite{gintis_social_2010,binmore1994economist}, given the right set of norms \cite{ostrom1990governing,bicchieri1999great,fehr_normative_2018}. (3) \emph{Intergenerational Norm Transmission}: We assess if norms can be sustained across generations of agents, and under what circumstances this occurs. (4) \emph{Norm Emergence and Convergence}: We investigate whether our assumption of shared normativity enables norms to organically emerge and stabilize through Bayesian learning without explicit sanctions.

In all of our experiments, we used a sample size of 13 runs for each condition, determined based on an effect size of $0.8$ with $\alpha = 0.05$ (one-sample $t$-test assuming a normal distribution).

\subsection{Passive Norm Learning}\label{sec:passive_learning}

In our norm learning experiments, we investigated whether a single learning agent $l$ could acquire the same norms believed in by a set of $M_{Exp} = 3$ experienced agents, each playing one of three roles $z \in Z := \{\textsf{C}, \textsf{F}, \textsf{E}\}$ (read as ``cleaner'', ``farmer'' and ``egalitarian'').
The learning agent also had an egalitarian role ($z_{Lea} = \textsf{E}$), and each episode lasted $H = 300$ steps. All experienced agents assigned a probability of 1 to the norms $N_\text{active} = \{\textsf{P1}, \textsf{P2}, \textsf{O1}, \textsf{O2}, \textsf{O3}\}$ shown in Figure~\ref{subfig:social_norms}. The full space $\mathcal{N}$ of 68 possible norms was generated by enumerating over a variety of predicates and numeric thresholds (see Appendix). The learning agent handled uncertainty over norms via thresholding, and had a high norm violation cost, motivating it to comply with learned norms.

\begin{figure}[t]
    \centering
    \includegraphics[width=\linewidth]{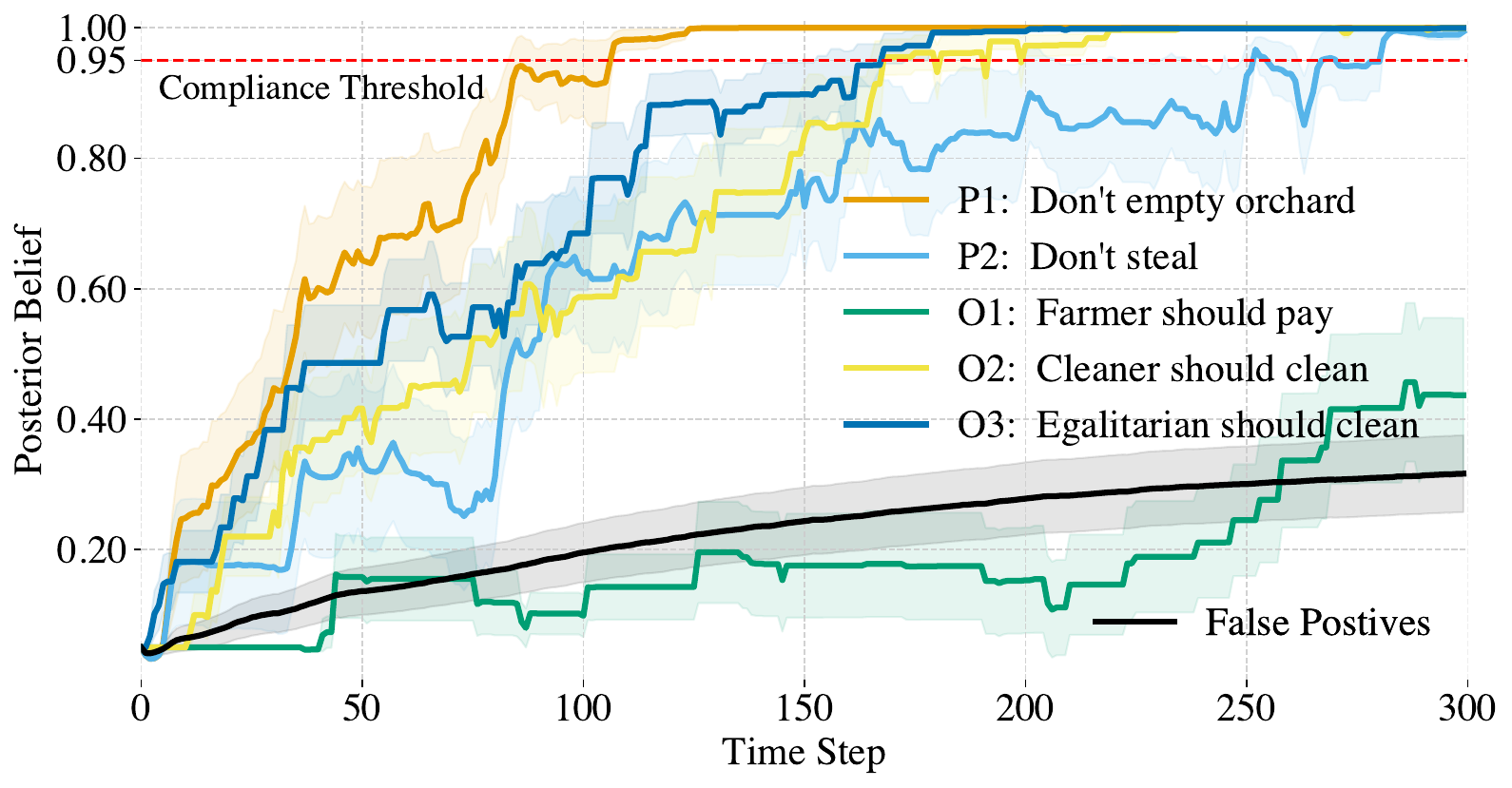}
    \caption{Passive Norm Learning. \textmd{Average posterior belief of the learner in each of the five norms practiced by the experienced agents, $N_\text{active}$. Most norms were acquired within $\leq 300$ steps.}}
    \label{fig:avg_rule_per_timestep}
   \vspace{-6pt}
\end{figure}

Figure~\ref{fig:avg_rule_per_timestep} demonstrates the versatility and efficiency of our learning model across a diverse set of norms. The majority of norms practiced by the experienced agents were rapidly and effectively internalized within 300 steps, in contrast to the \emph{millions} of steps required by model-free norm learning approaches \cite{koster_model-free_2020}. This was despite some norms being complied with only occasionally (e.g. the cleaner's obligation to clean the dirty river), providing a limited learning signal for observers. Interestingly, Obligation 1 (concerning the farmer's payment) was \emph{not} learned very well, with an average confidence of 0.4 even after 300 steps. This was likely due to the fact that the obligation was hard to satisfy, requiring the farmer to run after its assigned cleaner in order to pay them apples. As a result, the learning agent often inferred that the norm was \emph{non-existent} rather than merely hard to satisfy. To address this, more sophisticated norm learners should distinguish unintentional failure to comply with a norm from the absence of a norm. After all, many human obligations can be quite hard to fulfill, but we still take them to be in force despite the difficulty of fulfilling them.

\begin{table}[t]
    \centering
    \begin{tabular}{c|cccc}
    \toprule
     & $t/H$ = 25\% & 50\% & 75\% & 100\% \\
    $|N_\text{active}|$ & Pre/Rec & Pre/Rec & Pre/Rec & Pre/Rec \\
    \midrule
    0 & 0.00/0.00 & 0.00/0.00 & 0.00/0.00 & 0.00/0.00 \\
    1 & 0.15/0.09 & 0.08/0.40 & 0.06/0.54 & 0.06/0.57 \\
    2 & 0.19/0.27 & 0.14/0.59 & 0.13/0.76 & 0.12/0.82 \\
    3 & 0.24/0.37 & 0.21/0.63 & 0.18/0.77 & 0.18/\textbf{0.83} \\
    4 & 0.35/0.36 & 0.28/0.67 & 0.24/0.77 & 0.22/\textbf{0.83} \\
    5 & \textbf{0.47}/0.25 & 0.43/0.57 & 0.34/0.79 & 0.32/\textbf{0.83} \\
    \bottomrule
    \end{tabular}
    \vspace{4pt}
   \caption{\textmd{Average Precision (Pre) and Recall (Rec) for the learner successfully acquiring ($P(\nu) \geq 0.95$) the active norms $N_\text{active}$ after a certain \% of the episode, where $H$ denotes episode length.}}
   \label{tab:passive_learning}
   \vspace{-20pt}
\end{table}

Table~\ref{tab:passive_learning} shows our model's learning curve over time for varying numbers of actively practiced norms, as reflected in the precision and recall metrics. While we observed a higher recall with more time steps, precision was low in most cases, indicating that the learner was ``over-learning'' norms. From qualitative inspections of each run, we found that this was due to several reasons:
First, the learning agent sometimes captured the underlying dynamics of the environment. Even though norms were not explicitly followed, agents sometimes inferred them to be true if certain states of an environment were always avoided. For instance, agents learned to not be allowed to move when there was > 55\% of dirt in the river which was a state that was \emph{implicitly} avoided due to the river being cleaned on a regular basis. Such observations shed light on the phenomenon of overgeneralized norm learning in real-world contexts \cite{nichols2016rational}: Similar to unintentionally failed obligations, there exist ``accidentally fulfilled prohibitions'', and sophisticated learners should learn to recognize and account for both.

Second, we noticed that agents often learned \emph{logically overlapping} norms, reflecting the fact that some rules in $\mathcal{N}$ logically entailed other rules (e.g. if  $\textsf{Dirt(r) > 0.4}$ is true, then $\textsf{Dirt(r) > 0.3}$ is also true). This behavior was likely due to the mean-field approximation of the posterior in Equation \ref{eq:mean-field}, which prevented \emph{explaining away} among overlapping norms \cite{wellman1993explaining}. Future work should investigate how Bayesian norm learning can be made efficient without the mean-field approximation, perhaps via divide-and-conquer strategies \cite{lindsten2017divide} that first infer local posteriors over individual norms, then combine them into joint posteriors over sets of norms.

\subsection{Norm-Enabled Social Outcomes}

\begin{figure}[!t]
  \begin{subfigure}{0.9\linewidth}
    \centering
    \includegraphics[width=\linewidth]{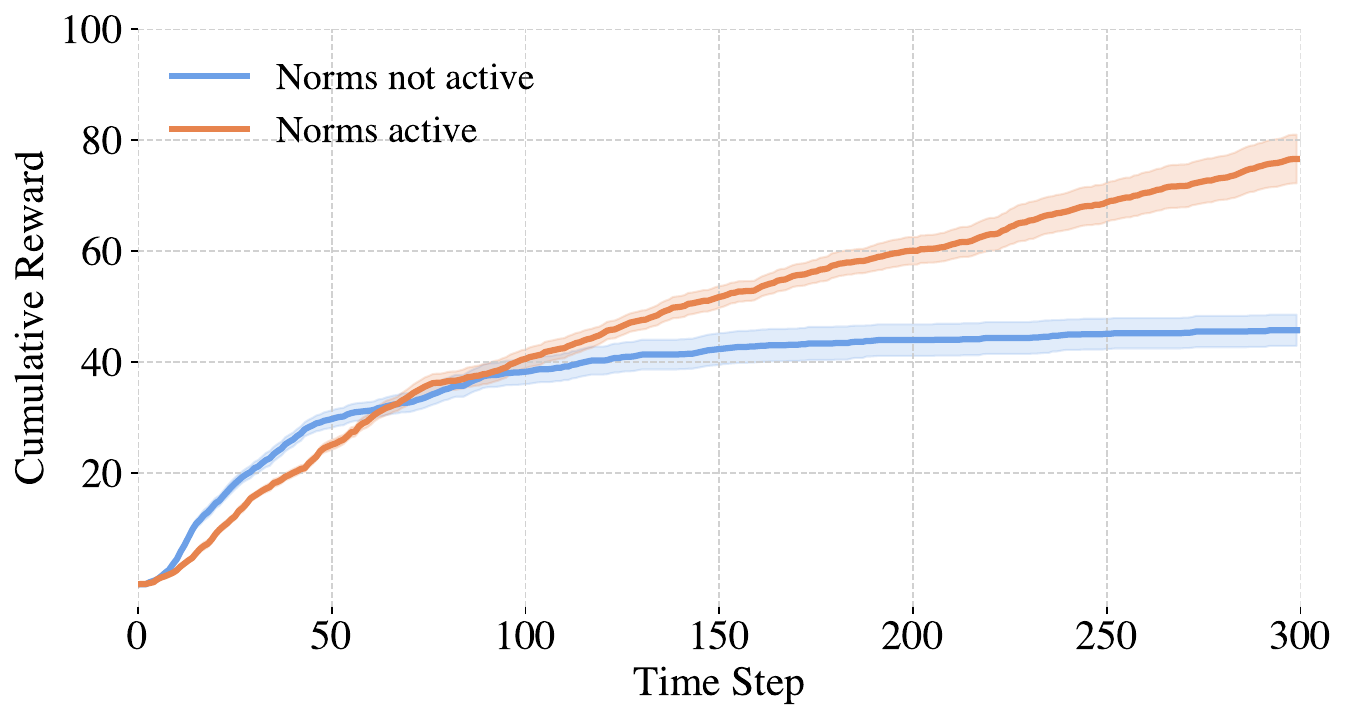}
    \subcaption{\textmd{Cumulative reward per set of practiced norms.}}
    \label{subfig:single_agent_welfare}
  \end{subfigure}  
  \begin{subfigure}{0.9\linewidth}
    \centering
    \includegraphics[width=\linewidth]{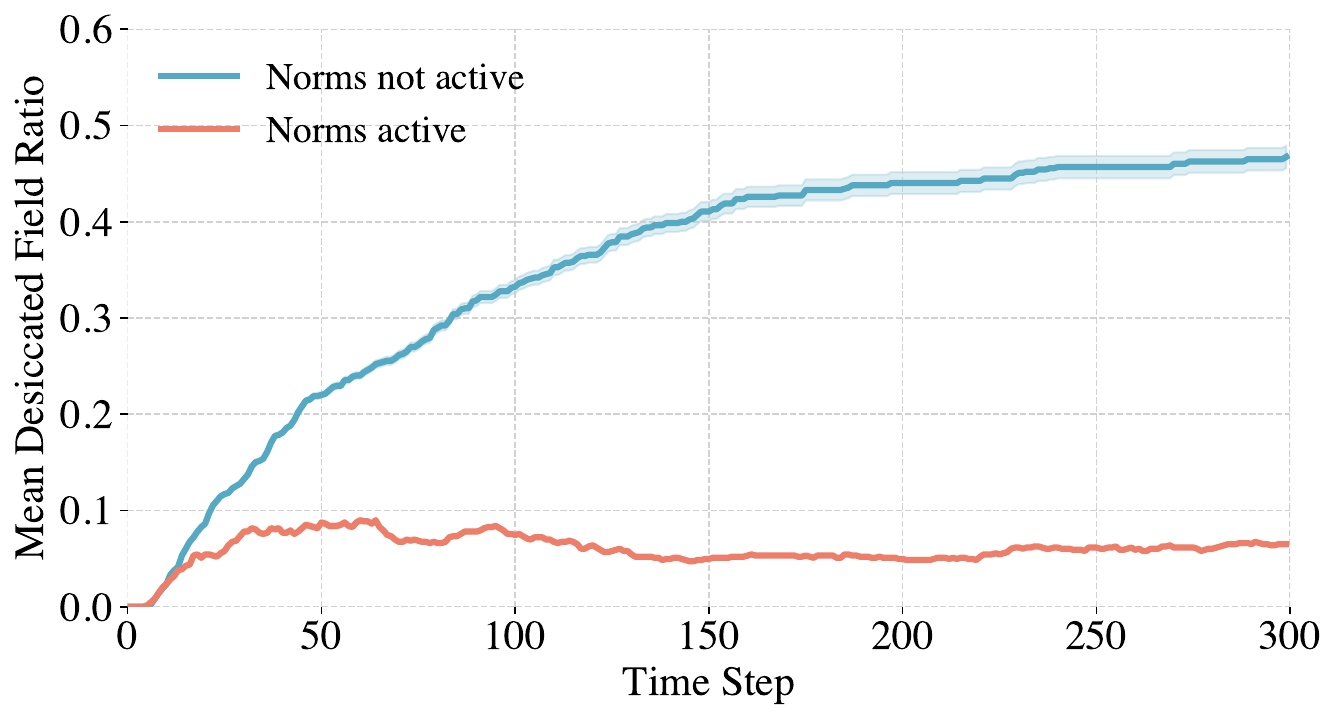}
    \subcaption{\textmd{Desiccated field ratio per set of practiced norms.}}
    \label{subfig:environment_destruction}
  \end{subfigure}
  \caption{Social outcomes \textmd{for two sets of norms $N$ practiced by the experienced agents: either all norms in $\{\textsf{P1}, \textsf{P2}, \textsf{O1}, \textsf{O2}, \textsf{O3}\}$ (\emph{Norms active}) or none of them (\emph{Norms not active}).}}
  \label{fig:welfare_graphs}
\end{figure}

Norms are instrumental in steering pro-social behavior and cultivating cooperation within societies \cite{scanlon_what_2000, hadfield_microfoundations_2014, fehr_normative_2018, dagostino_contemporary_2021}, and past research has illustrated the efficacy of norm compliance in addressing social challenges and enabling favorable outcomes \cite{ostrom1990governing,perolat_multi-agent_2017, lhaksmana_role-based_2018, hughes_inequity_2018, christoffersen_get_2022}. We investigated whether it was possible to achieve such outcomes with our norm learning approach, measuring social welfare in terms of the collective cumulative reward obtained by all agents \cite{vinitsky_learning_2023} (hereafter \emph{cumulative reward}) (Figure~\ref{subfig:single_agent_welfare}) and sustainability of the environment \cite{perolat_multi-agent_2017} (Figure~\ref{subfig:environment_destruction}), as measured by the ratio of desiccated apple fields in our environment relative to the total number of fields. While these were only two measures of desirable social outcomes (others might include equality, efficiency, etc. \cite{perolat_multi-agent_2017}), they nonetheless offered a useful measure of the benefit provided by norm-guided coordination. We conducted this analysis with the independent variable being the set $N$ of norms, where $N \in \{\emptyset, \{\textsf{P1}, \textsf{P2}, \textsf{O1}, \textsf{O2}, \textsf{O3}\} \}$. All other experimental conditions were consistent with those described earlier: one learning agent, and three experienced ones.

Perhaps unsurprisingly, the presence of norms \textsf{P1} through \textsf{O3} led to a clear increase in social welfare (Figure~\ref{fig:welfare_graphs}). Specifically, agents tended to achieve greater collective rewards with these norms in our environment (Figure~\ref{subfig:single_agent_welfare}). Unlike the scenario with no active norms, cumulative reward exhibited a consistent uptrend with time. Further evidence was provided by our sustainability metric: The percentage of desiccated fields (where apples do not regrow) increased rapidly in the absence of resource management norms, but remained at a low value when norms were present (Figure~\ref{subfig:environment_destruction}).

Of course, not all norms are equally beneficial, and some can even be harmful \cite{bicchieri1999great}. As Figures~\ref{fig:append_learner_rewards} and \ref{fig:append_mean_desiccated} in the Appendix show, some of the norms we considered had a weaker effect on social outcomes. In particular, the prohibitions against eating apples in less dense areas appeared to have limited impact. This might have been because the potential sustainability gain from obeying the prohibition directly traded off against the welfare gain of eating more apples. In contrast, once the cleaning obligations were fulfilled, the environment became significantly better maintained, enabling more apples to grow, and causing cumulative reward to keep increasing.

\subsection{Intergenerational Norm Transmission}\label{sec:intergen}

When agents have limited lifespans, or only join a community for fixed amounts of time, the stability of that community's norms requires the successful transmission of norms across generations \cite{aldous_social_1965, gonzalez_role_2021, tam_understanding_2015, hawkins_emergence_2019}. Since our account of norm learning depends on there being enough agents who \emph{practice} a norm for common knowledge of the norm to be maintained, we predicted that norms would be sustained as long there were enough experienced agents to learn from over a reasonable period of time \cite{trommsdorff_parentchild_2005, bowles_group_2006, watson_how_2016, bengtson_global_2018}. For shorter lifespans, we predicted that social norms would be unstable, drifting from those originally practiced.

\begin{figure}[!t]
  \begin{subfigure}{0.9\linewidth}
    \includegraphics[width=\linewidth]{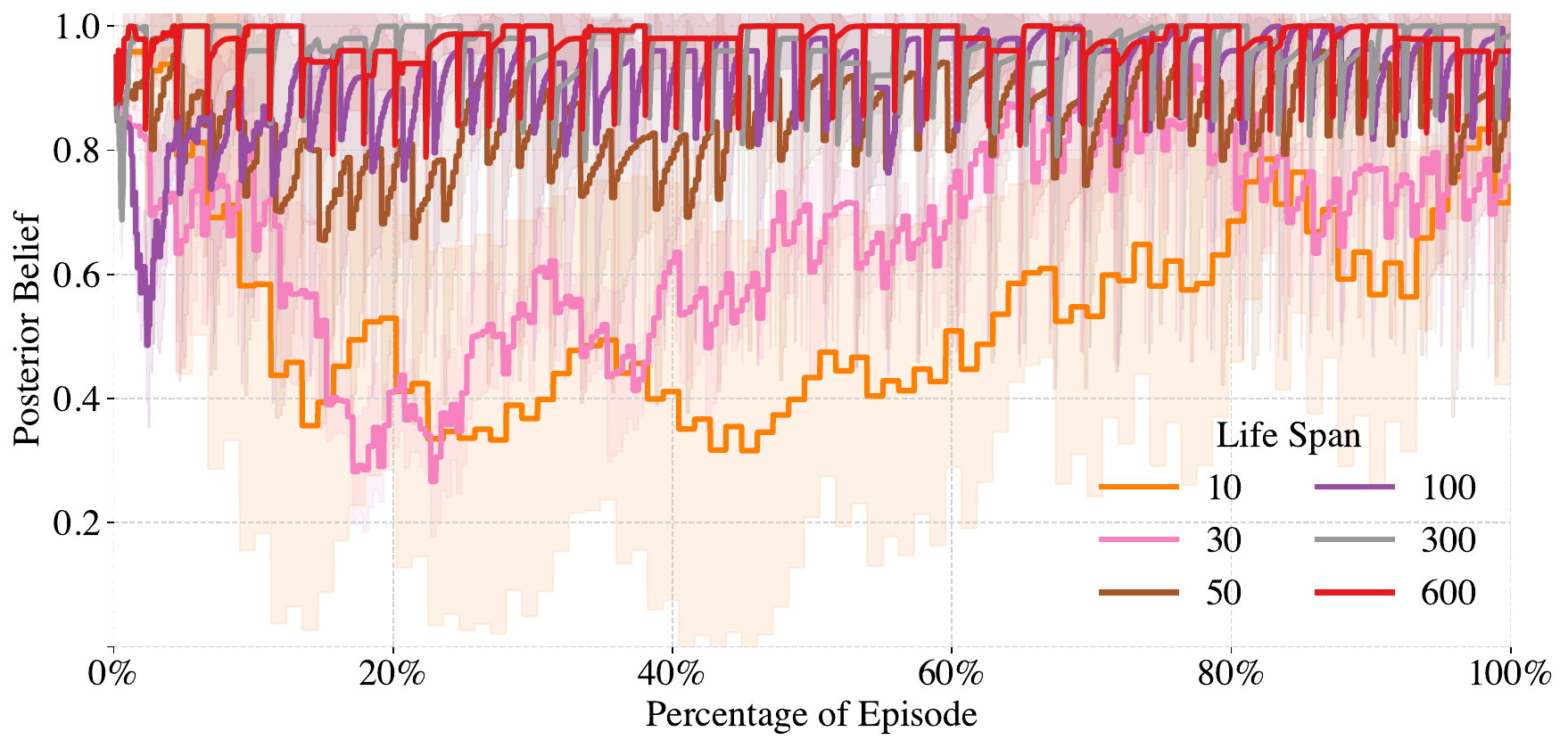}
    \subcaption{\textmd{P1: Don't pick up apples, if there are too few apples around.}}
    \label{subfig:norm_shift_prohibition}
  \end{subfigure}
  \begin{subfigure}{0.9\linewidth}
    \includegraphics[width=\linewidth]{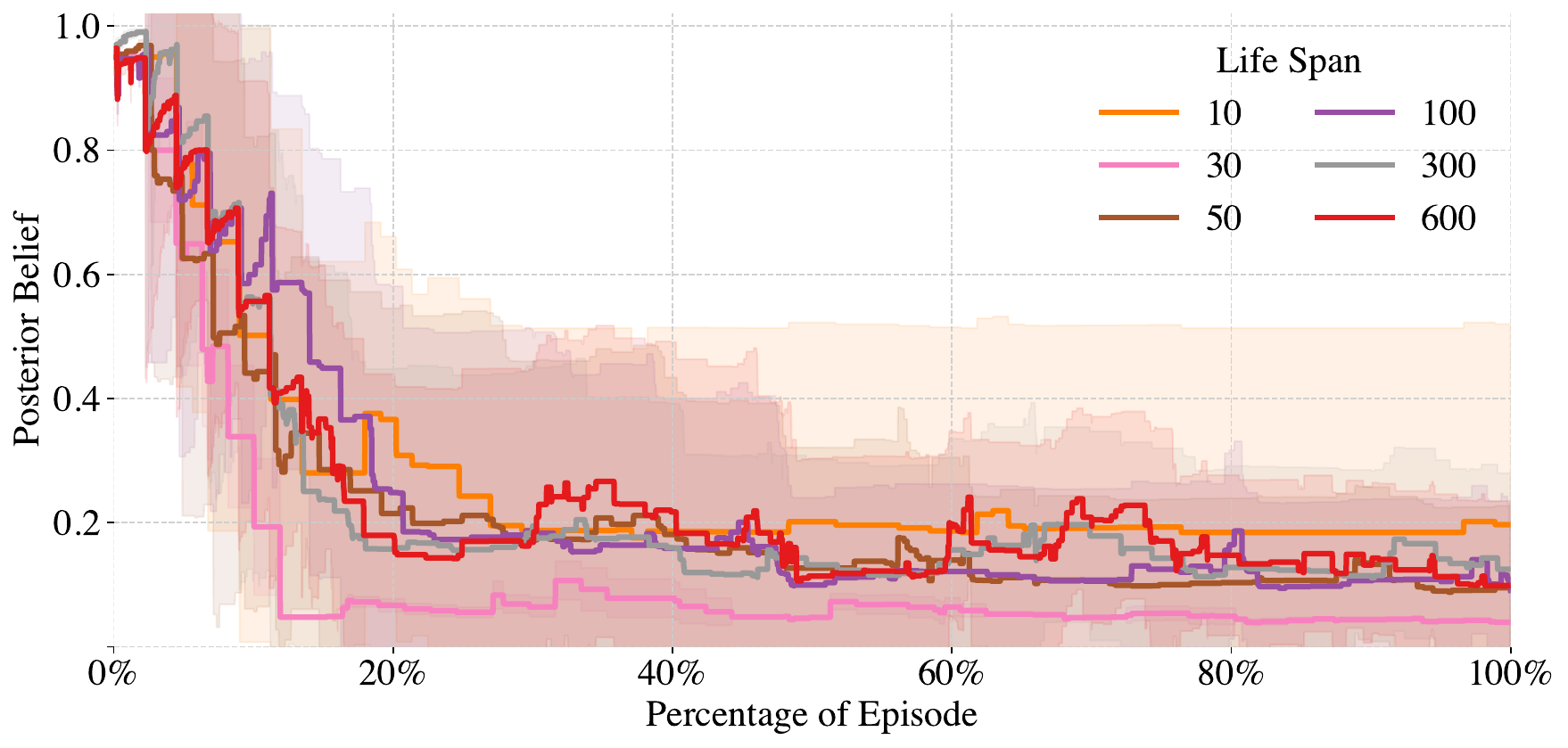}
    \subcaption{\textmd{O2: If you're a cleaner, clean the river when there is >30\% dirt.}}
    \label{subfig:norm_shift_obligation}
  \end{subfigure}  
  \begin{subfigure}{0.9\linewidth}
    \includegraphics[width=\linewidth]{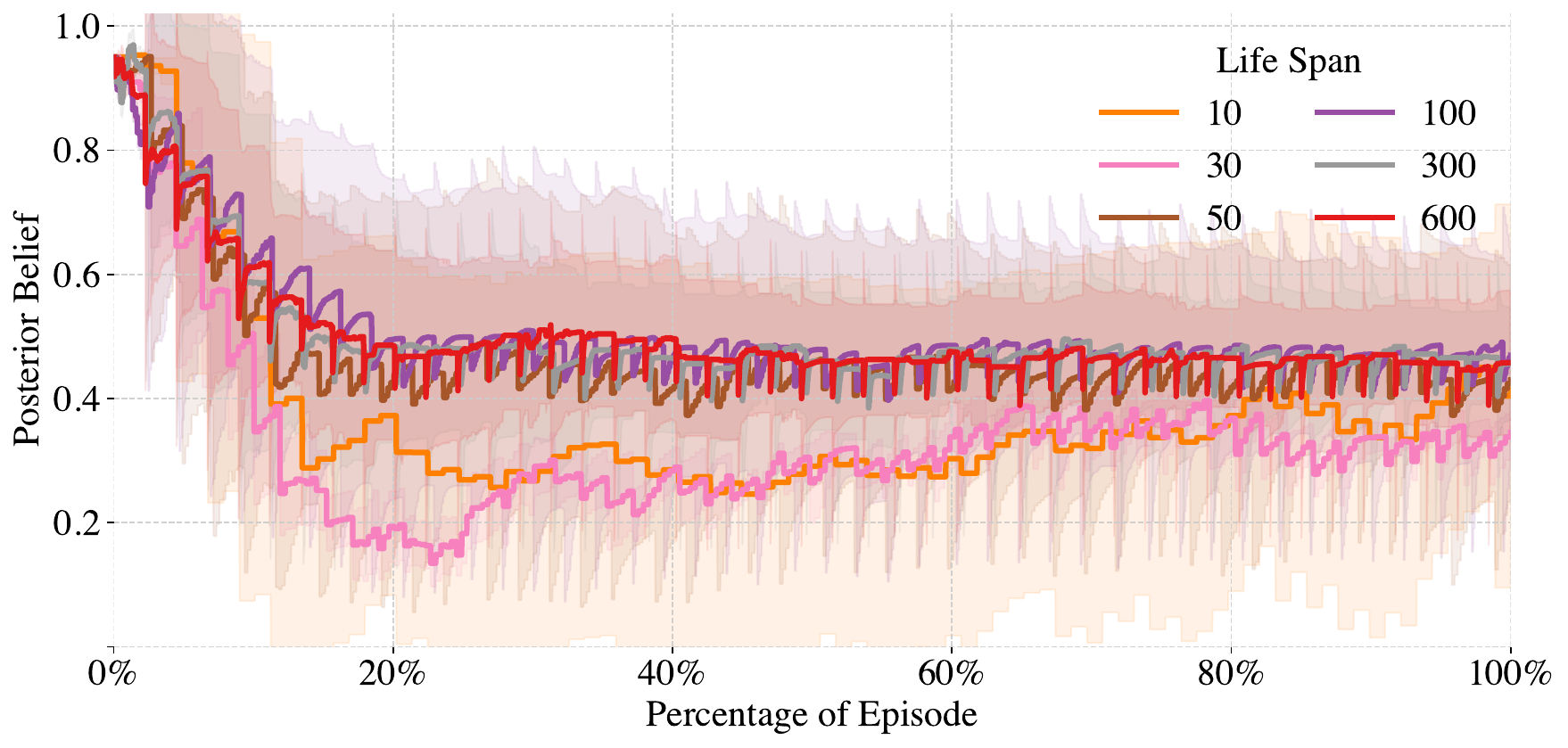}
    \subcaption{\textmd{Average for all five norms.}}
    \label{subfig:norm_shift_average}
  \end{subfigure}
  \caption{Intergenerational transmission of norms. \textmd{Mean norm belief averaged across all six agents across generations.}}
\label{fig:norm_drift_over_time}
\vspace{-20pt}
\end{figure}

To test these hypotheses, we ran an experiment with $M=6$ learning agents, while assigning each of the three role-associated appearances $z \in Z$ to two out of the six agents (effectively forming a parent-child pair for each role). Agents used sampling to select which norms to comply with. Each agent $i$ had a maximum life span of $L$, and an initial age of $\frac{i}{M} L$. Each run was simulated for $1.5 M L$ steps (i.e. three generational cycles). Upon an agent $i$'s demise, it was replaced by an equivalent ``child'' agent with the same role-associated appearance $z_i$, but with a low uniform prior over norms $P_i(\nu) = 1-\theta$ for all $\nu \in \mathcal{N}$. Initially, 3 experienced agents were present to seed the norms. As in earlier experiments, all agents were intrinsically motivated to comply with the learned norms.

The results of this experiment are shown in Figure \ref{fig:norm_drift_over_time}, revealing interesting patterns of norm transmission contingent upon the life spans of agents and the types of norms. Starting with the prohibition against eating too many apples (Figure~\ref{subfig:norm_shift_prohibition}), we observed striking rapidity in both its acquisition and transmission. Indeed, a life span as brief as 50 steps was sufficient for the effective transmission of these prohibition norms, aligning with the findings of our passive learning experiments (Figure~\ref{fig:avg_rule_per_timestep}). The amplification in learning speed can also be attributed to a larger number of agents to learn from (five instead of three), all adhering to the same prohibitive norm.

Obligation norms presented a more complex scenario (see e.g. Figure~\ref{subfig:norm_shift_obligation}). Due to the role-conditioned nature of the obligations we considered, learning opportunities were considerably more restricted compared to prohibitions. When learning the norms that applied to a particular role, an agent could only observe the two agents that were in that role, not all 6 agents. If they were the ``child'' agent in that role, they had only one agent to learn from (the ``parent'' in the same role). This resulted in a delayed learning curve for obligation norms, with agents requiring longer lifespans to learn and transmit them. Indeed, our results indicated that even an extended life span of 600 steps did not ensure that obligations were reliably transmitted.

Given the shorter learning time for obligations (300 steps) that we observed in Section \ref{sec:passive_learning}, this inability to transmit obligations was unexpected. What we found instead was collective forgetting of all initially active obligations. Beyond the fact that obligations were less frequently observed, a possible factor was sampling: Since agents resampled the norms they would comply with every $K = 10$ steps, this added stochasticity that interfered with the long-horizon goals involved in obligation-oriented planning: If an obligation was not sampled at any step while fulfilling its goal, an agent would give up on the goal and go back to maximizing reward. This in turn created less of a learning signal for other agents.

Notwithstanding these unanticipated factors, our overall hypothesis about norm transmission was validated: Under \emph{some} conditions, social norms are successfully maintained and transmitted (Figure \ref{subfig:norm_shift_average}), and this is more likely to occur with more opportunities to learn those norms from agents who practice them. While more theoretical work is required, this suggests that in some regimes, norms are not only in a normative equilibrium but \emph{evolutionarily stable} \cite{smith1982evolution}. In contrast, when observations are scarce, norms might be abandoned \cite{panke2012international}, even by agents who used to comply with them.

\subsection{Norm Emergence and Convergence}\label{sec:emergence}

\begin{figure}[!t]
\centering
    \includegraphics[width=\linewidth]{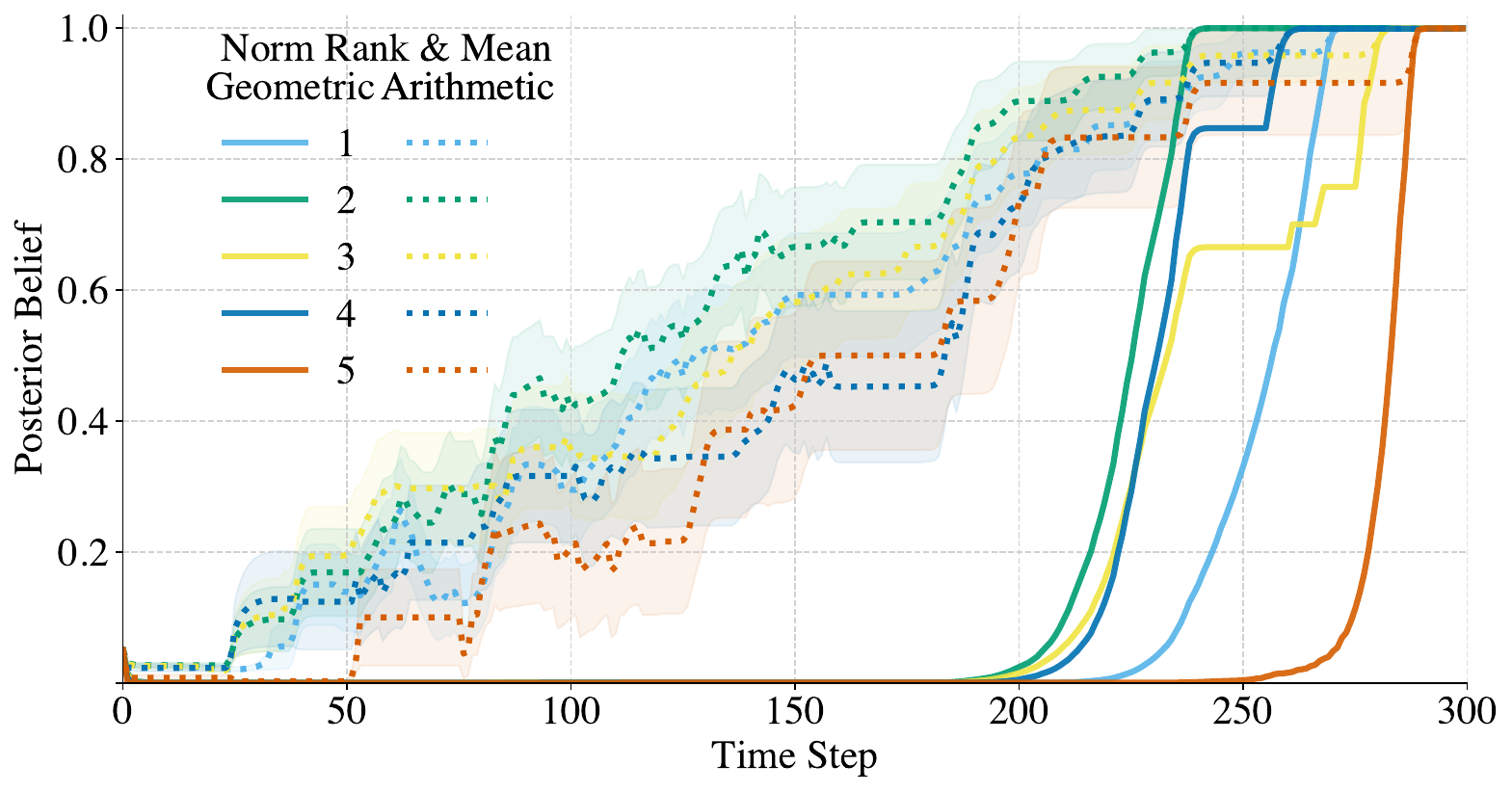}
    \caption{Norm emergence and convergence. \textmd{Geometric and arithmetic mean posterior beliefs of all agents for the top 5 norms with final belief $P^{t=300}(n)\geq \theta = 0.95$.}}
\label{fig:convergence_rate}
\end{figure}

The emergence of norms is a complex phenomenon that has been studied both empirically \cite{axelrod1981evolution,savarimuthu_norm_2009, shibusawa_norm_2014,vinitsky_learning_2023} and theoretically \cite{ullmann2015emergence, hawkins_emergence_2019,morris-martin_norm_2019}. While prior work has studied norm emergence through adaption of policies in response to punishment \cite{savarimuthu_norm_2009}, or learning to punish in accordance with public signals \cite{vinitsky_learning_2023}, our approach builds upon models of ad-hoc cooperation through joint intentionality \cite{wu_too_2021}, bootstrapping a shared set of norms in the same way that a team of cooperating agents can bootstrap a common goal without explicit communication \cite{tang2020bootstrapping}. Unlike the aforementioned mechanisms, this model of norm emergence does not require enforcement, as long as agents are sufficiently motivated (intrinsically or extrinsically) to comply with emergent norms.

To test whether such bootstrapping in fact occurs, we simulated a population of $M=3$ learning agents with $z \in \{\textsf{C}, \textsf{E}, \textsf{F}\}$ (every role represented once). As in our norm transmission experiments, agents select norms to comply with via sampling, effectively performing a form of \emph{norm exploration} when uncertain about which norms hold.

Figure~\ref{fig:convergence_rate} summarizes the results of these experiments and demonstrates the emergence of shared norms among our agents. The gradual increase in the arithmetic mean over multiple runs signifies an emergent belief in the existence of certain norms, even though such beliefs may not initially be shared by all agents. This eventually results in a sharp rise in the geometric mean, demonstrating convergence and unanimous agreement among agents on specific norms. Through repeated interaction and learning, our agents coalesce around a set of norms they collectively perceive as true, illustrating how pure observational learning combined with experimental norm compliance can eventually lead to a stable set of norms.

Examining the emergent norms more closely (Table~\ref{tab:append_emerging_norms} in the Appendix), we found that many of the learned norms were non-functional, or ``silly'' norms \cite{hadfield-menell_legible_2018}. This was likely due to intrinsic norm violation costs, which help to stabilize inefficient or even harmful norms \cite{bicchieri_social_2018}. As such, we expect that emergent norms are more likely to be beneficial when agents are more strategic about which norms they comply with. Nonetheless, recent work has shown how non-functional norms can play an epistemic role in stabilizing a normative system \cite{hadfield-menell_legible_2018}, and our model provides one mechanism for how such auxiliary norms might arise.

\section{Discussion and Future Work}\label{sec:discussion}
How should we build autonomous systems that learn to comply with the norms of their social environments? In this paper, we introduced a Bayesian framework for decentralized norm learning in long-horizon multi-agent settings, formalized as \emph{norm-augmented Markov games} (NMGs). In these games, each agent pursues their own objectives while complying with the norms that they infer from collective behavior. In doing so, we show how agents can \emph{rapidly acquire structured norms} by observing experienced agents, \emph{sustain common knowledge} of these norms in a changing population, and \emph{converge upon shared norms} even when none are initially present. These features not only explain important facets of human normativity, but may also prove crucial for autonomous agents.

Nonetheless, many conceptual and technical questions about norm learning remain unexplored by our experiments, or are not explicitly accounted for by our modeling framework. We review these questions in the following discussion, and consider how our framework could be extended to address them.

\paragraph{The Role of Sanctions in Learning and Sustaining Norms} Perhaps the most important difference between our experiments and previous work on decentralized norm learning \cite{axelrod1981evolution,savarimuthu_norm_2009,vinitsky_learning_2023} is the absence of \emph{sanctions} --- acts of punishment or approval --- in motivating and sustaining norm compliance. Rather, our experiments focused on the \emph{epistemic} aspect of norm learning, i.e., learning the \emph{content} of normative rules, and ensuring they remain commonly known, leaving open the \emph{motivational} aspect of norm learning. In part, this is because motivational learning is not as important for machines: We can often design them to \emph{want} to comply with norms.

That said, nothing about the NMG formalism precludes the study of sanctions, which are important when we cannot control the motivations of other agents. Sanctioning could be introduced as a primitive action (e.g. the ``zapping beams'' supported by  Melting Pot \cite{leibo_scalable_2021,agapiou_melting_2023}), a learned policy (e.g. taking others' apples in retaliation for theft), or a (meta)-norm that obligates punishment of agents who do not comply with norms. Sophisticated agents might then learn or plan to use sanctions against norm violators. Would-be violators --- e.g. agents with no intrinsic violation costs --- might learn to expect punishment for norm violation, thereby choosing to comply. In fact, by separating the epistemic and motivational aspects of norm learning, NMGs may deliver a clearer account of the role of sanctions: Sanctions provide both \emph{evidence} that a norm exists, and \emph{motivation} to comply with it. By adjusting their strengths, we can investigate their contributions to norm learning and stability. We could also disambiguate intrinsic motivation accounts of punishment \cite{darley2000incapacitation} from more strategic accounts, where agents punish to signal support for existing norms \cite{hadfield_what_2012,cushman2019punishment,radkani2022modeling}, influencing the normative beliefs of other agents in ways that they deem beneficial.

\paragraph{Model-Free vs. Model-Based Norm Learning}
Our paper takes a model-based approach to norm-oriented learning and planning, achieving orders of magnitude more sample efficiency than primarily model-free approaches \cite{koster_model-free_2020,vinitsky_learning_2023}. However, this comes at the cost of greater runtime complexity. To address these challenges, model-free and model-based approaches can be fruitfully combined. For example Bayesian learning can be accelerated via bottom-up proposals \cite{zhu2000integrating} and amortized inference \cite{wang2018meta,dasgupta2020theory}. 

The question of model-free vs. model-based learning also relates to how norms are \emph{represented}: Some norms --- those that we tend to call \emph{practices} or \emph{rituals} \cite{hagen2010propriety} -- are perhaps best represented as \emph{habits} \cite{pauli2018computational} or \emph{skills}, which can be learned model-free \cite{koster_model-free_2020}. Other norms are more naturally thought of as \emph{abstract rules}, classifying behavior by its permissibility \cite{von1981logic}, while enabling public interpretability and generalization to new scenarios \cite{hadfield_microfoundations_2014}. Our framework emphasizes the latter because we believe it is important for the conceptual richness of normative cognition: If norms lacked a symbolic, language-like structure, we would find it hard to compose, generalize, communicate, and argue about them \cite{mikhail2007universal}.

While symbolic rules have become less popular in machine learning, recent advances in large language models (LLMs) suggest that they can made considerably more scalable. As methods like Constitutional AI \cite{bai2022constitutional} suggest, a norm or principle $\nu$ could be represented in natural language itself, using an LLM to \emph{classify} whether behavior complies with the norm \cite{nay_law_2022}. Alternatively, LLMs might \emph{translate} natural language, such as legal documents, into formal specifications \cite{camilleri_contracts_2017,liu2023lang2ltl}, enabling grounded and reliable reasoning in a particular domain \cite{wong_word_2023,olausson2023linc,zhixuan2024pragmatic}. LLMs could thus serve as priors over norms $P_i^0(N)$ for a norm learner, or as approximate posteriors $P_i(N | u)$ over norms $N$ given a piece of normative language $u$.

Symbolic norms do not eliminate the value of model-free learning, however: In general, learning how to \emph{generate} norm-compliant behavior is harder than learning to \emph{discriminate} it \cite{wu2019tale}. Oftentimes, it may be more efficient to learn habitual practices that comply with a norm, rather than planning them. Relatedly, the \emph{motivation} to comply with norms might be learned via model-free mechanisms \cite{cushman2013action}, even if norms are learned in a model-based manner.

\paragraph{Normative Reasoning and Norm Adaptation}

In our experiments, agents passively learn existing norms, then comply with them once learned. Yet humans are significantly more flexible. We do not just learn norms, but also decide when it makes sense to follow them \cite{levine2020logic,kwon2023not}, and adjudicate when they are fairly applied \cite{hadfield2017rules}. This is because we can \emph{reason} about the function of norms \cite{kohlberg1977moral} and how they affect each of us \cite{scanlon_what_2000}, adapting them to novel circumstances \cite{morris-martin_norm_2019}. Yet given the complexity of reasoning about ideal norms, it is costly to do this all the time. Instead, if we are to build normatively competent agents, we should perhaps take a leaf from recent accounts of how humans manage this complexity \cite{levine_resource-rational_2023}: defaulting to learned rules in tried-and-true situations, adjusting them in new scenarios, and forging new norms via agreement when it is feasible and worthwhile. By shifting fluidly between these levels of normative cognition, we might eventually build systems that not only \emph{comply} with the norms of our society, but help us deliberate upon and \emph{improve} them.

\section*{Acknowledgments}

We thank our anonymous reviewers for their thoughtful feedback as well as the Summer Research Fellowship for Principles of Intelligent Behavior in Biological and Social Systems (PIBBSS) for funding parts of this project. Tan Zhi-Xuan is supported by the Open Philanthropy AI Fellowship.

\section*{Supplementary Material}

Source code is available at \href{https://github.com/ninell-oldenburg/social-contracts}{\emph{https://github.com/ninell-oldenburg/social-contracts}}.

\bibliographystyle{ACM-Reference-Format}
\balance
\bibliography{Bibliographies/refs,Bibliographies/extra}

\clearpage
\onecolumn
\appendix
\section*{Appendix}\label{sec:appendix}
\renewcommand{\thefigure}{A\arabic{figure}}
\setcounter{figure}{0}

\renewcommand{\thetable}{A\arabic{table}}
\setcounter{table}{0} 

\renewcommand{\thesubsection}{A\arabic{subsection}}
\setcounter{subsection}{0} 

\subsection{Model and Environment Parameters}
\begin{figure}[!h]
    \centering
    \includegraphics[width=14cm]{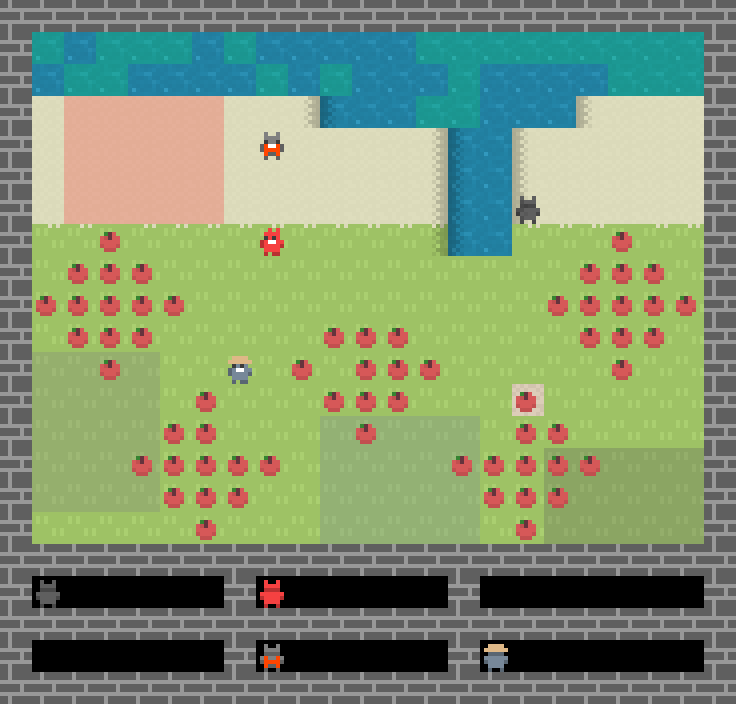}
  \caption{Example frame of our environment. \textmd{The three experienced agents are depicted in grey each representing a specific role based on their appearance. The learning agent is completely colored, mimicking the ``egalitarian" role based on its appearance. Each agent's territory is visible as an agent-colored overlay. The river is located on the top of the frame with brighter parts representing the dirt. Apples are located in orchards that might, in turn, be in part on an agent's territory. If a field is desiccated and hence unlikely but not impossible to regrow apples, it is signified by a sand-like color underneath.}}
  \label{fig:environment}
\end{figure}

\paragraph{Parameters for the Norm-augmented Markov Game.}

The individual reward for eating an apple for every agent is $1$ while the action cost is $0.01$. Each prohibition norm violation results in a violation cost of $1$ that is applied to the value function as described in Equation~\ref{eq:value-prohib} while, as mentioned above, the reward for the fulfillment of an obligation is $1$, as specified in Equation~\ref{eq:value-oblig}. Our specific environment's prohibition and obligation norms are specified in Figure~\ref{subfig:social_norms}.

\paragraph{Parameters for the Norm-compliant Planning Algorithm.}

We use real-time dynamic programming (RTDP) \cite{barto_learning_1995} for the norm-compliant planning algorithm with a rollout depth of $20$ every $2$ steps and a discount factor of $\gamma = 0.9$. Agents thereby comply with the norms at threshold $\theta = 0.95$ or a sampling frequency of $K = 10$ for the intergenerational norm learning setting (Section~\ref{sec:intergen}) and $K = 15$ for the norm emergence setting (Section~\ref{sec:emergence}) to secure more robust emergence possibilities.

\paragraph{Parameters for the Norm-learning Algorithm.}

Every potential norm out of the 68 generated ones (see Table~\ref{tab:append_potential_norms}) was given an initial prior of $P(\nu) = 0.05$. We employed a Boltzmann temperature of $1$.

\newpage
\begin{longtable}{rl}
\caption{Generated list of feasible and non-trivial potential norms $\mathcal{N}$ \textmd{up to a length of three parameters. (e.g. norms 24-31). The parameters are variables of the agents' observation space. The norms are feasible in the sense that the state the parameters specify is reachable and non-trivial in the sense that the state is non-tautological.}} \\
\label{tab:append_potential_norms}
\textbf{Prohibitions} & \\
1 & If a cell holds an apple $\to$ Don't move there.\\

2 & If a cell is not your property $\to$ Don't move there.\\

3 & If the river dirt fraction > 0.3 $\to$ Don't move at all.\\

4 & If the river dirt fraction > 0.35 $\to$ Don't move at all.\\

5 & If the river dirt fraction > 0.4 $\to$ Don't move at all.\\

6 & If the river dirt fraction > 0.45 $\to$ Don't move at all.\\

7 & If the river dirt fraction > 0.5 $\to$ Don't move at all.\\

8 & If the river dirt fraction > 0.55 $\to$ Don't move at all.\\

9 & If the river dirt fraction > 0.6 $\to$ Don't move at all.\\

10 & If you're looking towards North $\to$ Don't move at all.\\

11 & If you're looking towards East $\to$ Don't move at all.\\

12 & If you're looking towards South $\to$ Don't move at all.\\

13 & If you're looking towards West $\to$ Don't move at all.\\

14 & If a cell holds an apple and it's not your property $\to$ Don't move there.\\

15 & If a cell holds an apple and the number of apples around < 1 $\to$ Don't move there.\\

16 & If a cell holds an apple and the number of apples around < 2 $\to$ Don't move there.\\

17 & If a cell holds an apple and the number of apples around < 3 $\to$ Don't move there.\\

18 & If a cell holds an apple and the number of apples around < 4 $\to$ Don't move there.\\

19 & If a cell holds an apple and the number of apples around < 5 $\to$ Don't move there.\\

20 & If a cell holds an apple and the number of apples around < 6 $\to$ Don't move there.\\

21 & If a cell holds an apple and the number of apples around < 7 $\to$ Don't move there.\\

22 & If a cell holds an apple and the number of apples around < 8 $\to$ Don't move there.\\

23 & If a cell holds an apple and the number of apples around < 8 $\to$ Don't move there.\\

24 & If a cell holds an apple and it's not your property and the number of apples around < 1 $\to$ Don't move there.\\

25 & If a cell holds an apple and it's not your property and the number of apples around < 2 $\to$ Don't move there.\\

26 & If a cell holds an apple and it's not your property and the number of apples around < 3 $\to$ Don't move there.\\

27 & If a cell holds an apple and it's not your property and the number of apples around < 4 $\to$ Don't move there.\\

28 & If a cell holds an apple and it's not your property and the number of apples around < 5 $\to$ Don't move there.\\

29 & If a cell holds an apple and it's not your property and the number of apples around < 6 $\to$ Don't move there.\\

30 & If a cell holds an apple and it's not your property and the number of apples around < 7 $\to$ Don't move there.\\

31 & If a cell holds an apple and it's not your property and the number of apples around < 8 $\to$ Don't move there.\\
\\

\textbf{Obligations} & \\
32 & If the river dirt fraction > 0.3 and you look like a cleaner $\to$ Clean the river.\\

33 & If the river dirt fraction > 0.3 and you look like a farmer $\to$ Clean the river.\\

34 & If the river dirt fraction > 0.3 and you look like an egalitarian $\to$ Clean the river.\\

35 & If the river dirt fraction > 0.35 and you look like a cleaner $\to$ Clean the river.\\

36 & If the river dirt fraction > 0.35 and you look like a farmer $\to$ Clean the river.\\

37 & If the river dirt fraction > 0.35 and you look like an egalitarian $\to$ Clean the river.\\

38 & If the river dirt fraction > 0.4 and you look like a cleaner $\to$ Clean the river.\\

39 & If the river dirt fraction > 0.4 and you look like a farmer $\to$ Clean the river.\\

40 & If the river dirt fraction > 0.4 and you look like an egalitarian $\to$ Clean the river.\\

41 & If the river dirt fraction > 0.45 and you look like a cleaner $\to$ Clean the river.\\

42 & If the river dirt fraction > 0.45 and you look like a farmer $\to$ Clean the river.\\

43 & If the river dirt fraction > 0.45 and you look like an egalitarian $\to$ Clean the river.\\

44 & If the river dirt fraction > 0.5 and you look like a cleaner $\to$ Clean the river.\\

45 & If the river dirt fraction > 0.5 and you look like a farmer $\to$ Clean the river.\\

46 & If the river dirt fraction > 0.5 and you look like an egalitarian $\to$ Clean the river.\\

47 & If the river dirt fraction > 0.55 and you look like a cleaner $\to$ Clean the river.\\

48 & If the river dirt fraction > 0.55 and you look like a farmer $\to$ Clean the river.\\

49 & If the river dirt fraction > 0.55 and you look like an egalitarian $\to$ Clean the river.\\

50 & If the river dirt fraction > 0.6 and you look like a cleaner $\to$ Clean the river.\\

51 & If the river dirt fraction > 0.6 and you look like a farmer $\to$ Clean the river.\\

52 & If the river dirt fraction > 0.6 and you look like an egalitarian $\to$ Clean the river.\\

53 & If the number of steps you haven't paid anyone > 10 and you look like a cleaner $\to$ Pay someone.\\

54 & If the number of steps you haven't paid anyone > 10 and you look like a farmer $\to$ Pay someone.\\

55 & If the number of steps you haven't paid anyone > 10 and you look like an egalitarian $\to$ Pay someone.\\

56 & If the number of steps you haven't paid anyone > 15 and you look like a cleaner $\to$ Pay someone.\\

57 & If the number of steps you haven't paid anyone > 15 and you look like a farmer $\to$ Pay someone.\\

58 & If the number of steps you haven't paid anyone > 15 and you look like an egalitarian $\to$ Pay someone.\\

59 & If the number of steps you haven't paid anyone > 20 and you look like a cleaner $\to$ Pay someone.\\

60 & If the number of steps you haven't paid anyone > 20 and you look like a farmer $\to$ Pay someone.\\

61 & If the number of steps you haven't paid anyone > 20 and you look like an egalitarian $\to$ Pay someone.\\

62 & If the number of steps you haven't paid anyone > 25 and you look like a cleaner $\to$ Pay someone.\\

63 & If the number of steps you haven't paid anyone > 25 and you look like a farmer $\to$ Pay someone.\\

64 & If the number of steps you haven't paid anyone > 25 and you look like an egalitarian $\to$ Pay someone.\\

65 & If the number of steps you haven't paid anyone > 30 and you look like a cleaner $\to$ Pay someone.\\

66 & If the number of steps you haven't paid anyone > 30 and you look like a farmer $\to$ Pay someone.\\

67 & If the number of steps you haven't paid anyone > 30 and you look like an egalitarian $\to$ Pay someone.\\

68 & If another agent violated a norms $\to$ Sanction them.\\
\end{longtable}

\clearpage
\subsection{Additional Results}

\begin{figure}[!h]
  \begin{subfigure}{0.45\linewidth}
    \centering
    \includegraphics[width=\linewidth]{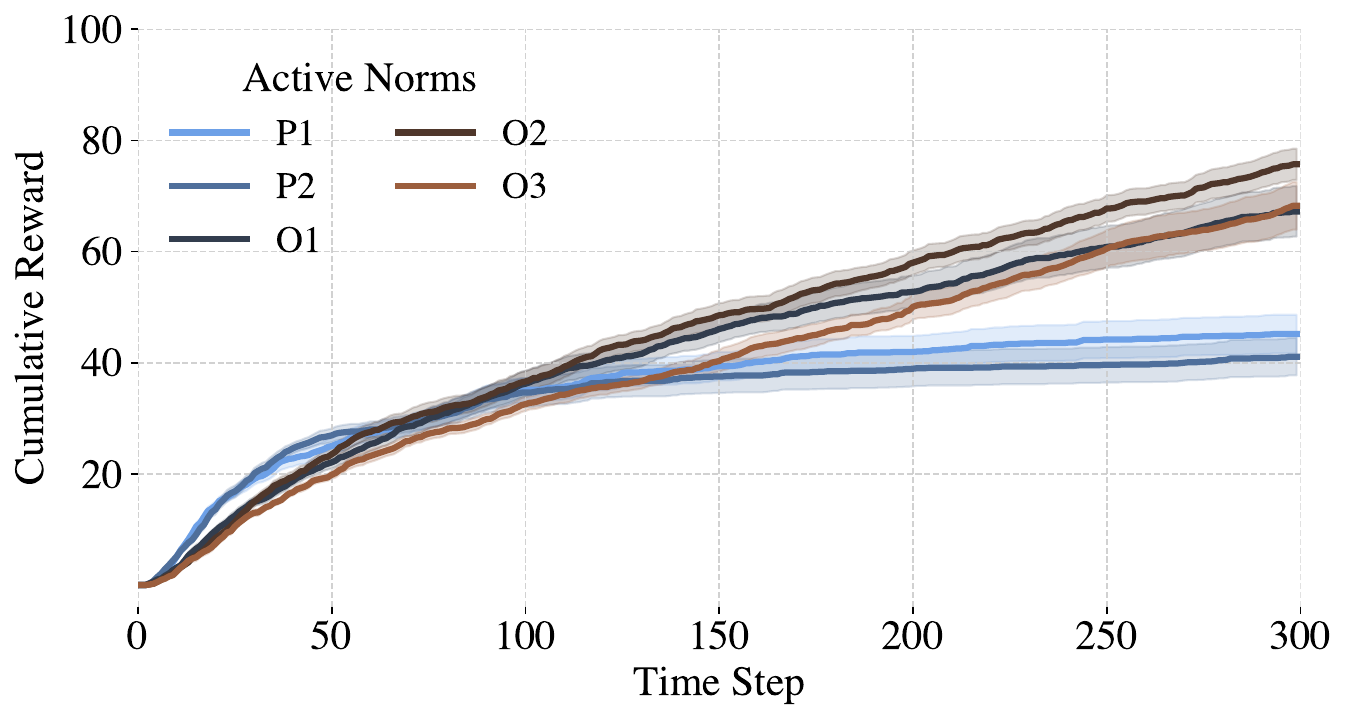}
    \subcaption{\textmd{One norm.}}
    \label{subfig:append_one_rule_desiccated}
  \end{subfigure}  
  \begin{subfigure}{0.45\linewidth}
    \centering
    \includegraphics[width=\linewidth]{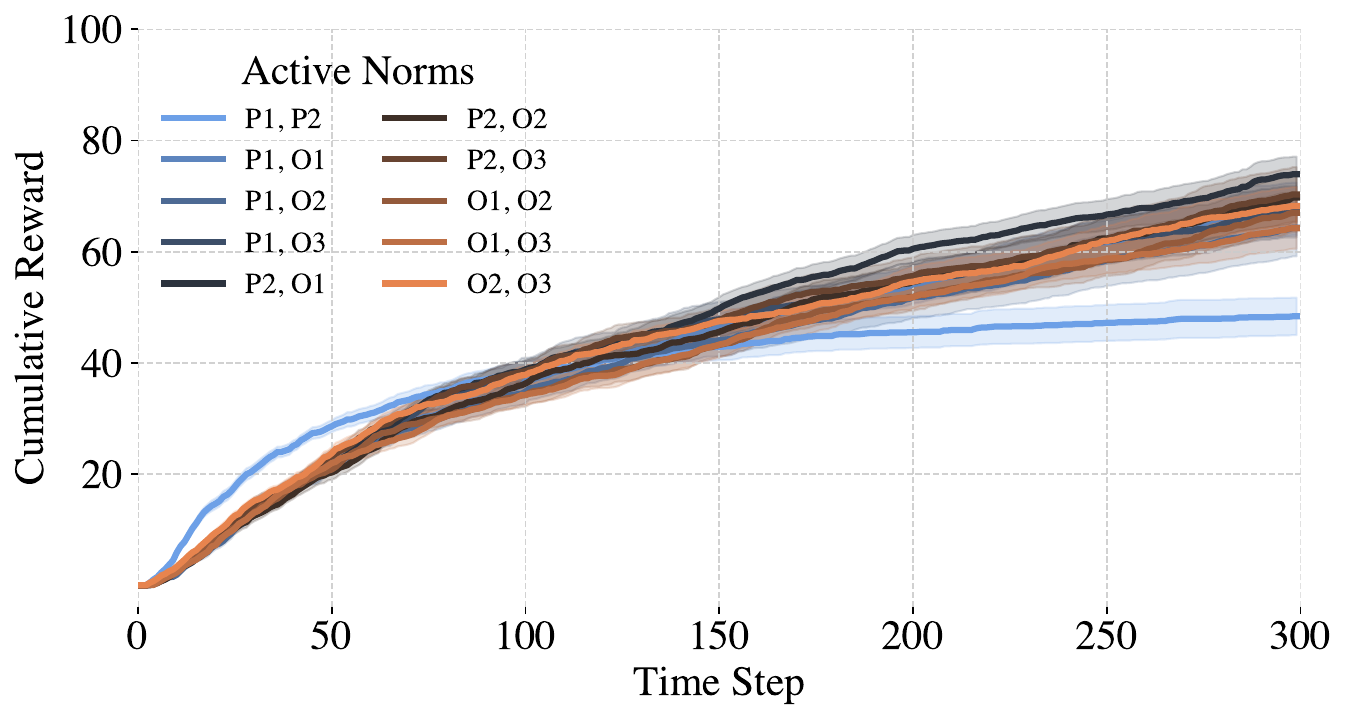}
    \subcaption{\textmd{Two norms combinations.}}
    \label{subfig:append_two_rule_desiccated}
  \end{subfigure}  
  \begin{subfigure}{0.45\linewidth}
    \centering
    \includegraphics[width=\linewidth]{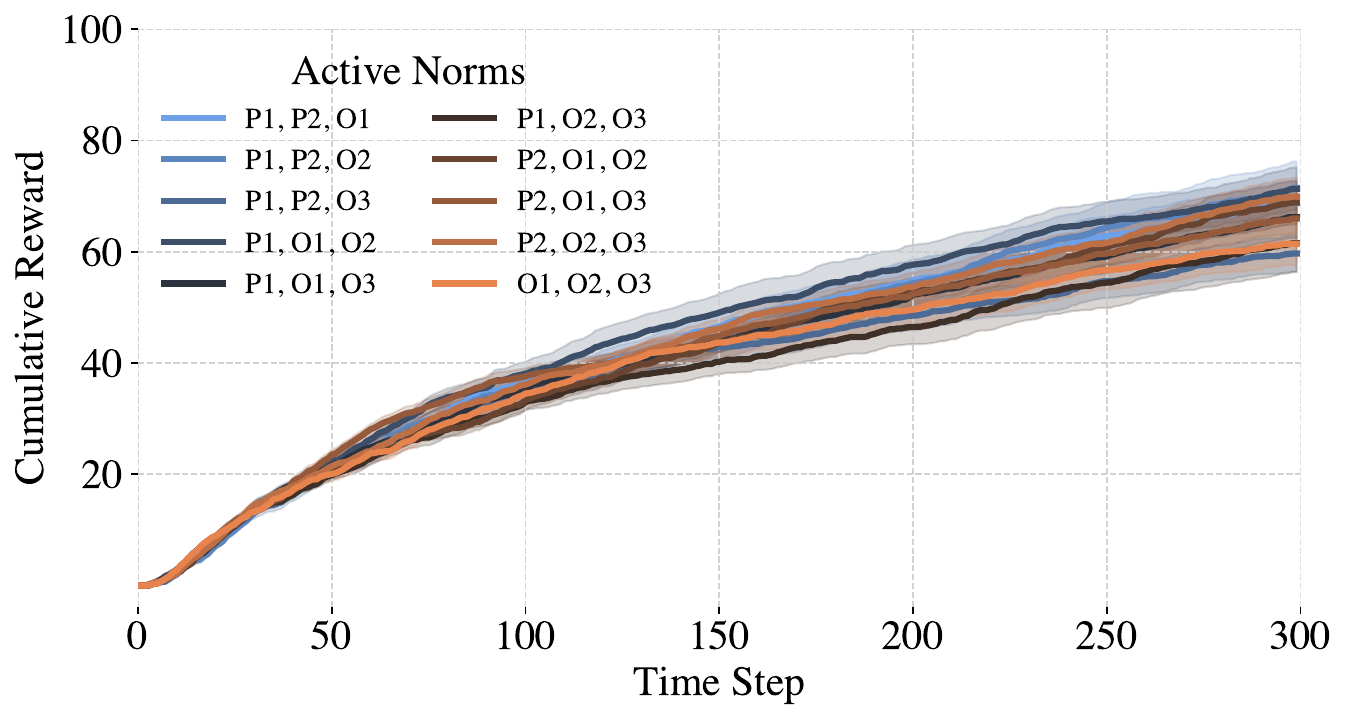}
    \subcaption{\textmd{Three norms combinations.}}
    \label{subfig:append_three_rule_desiccated}
  \end{subfigure}  
  \begin{subfigure}{0.45\linewidth}
    \centering
    \includegraphics[width=\linewidth]{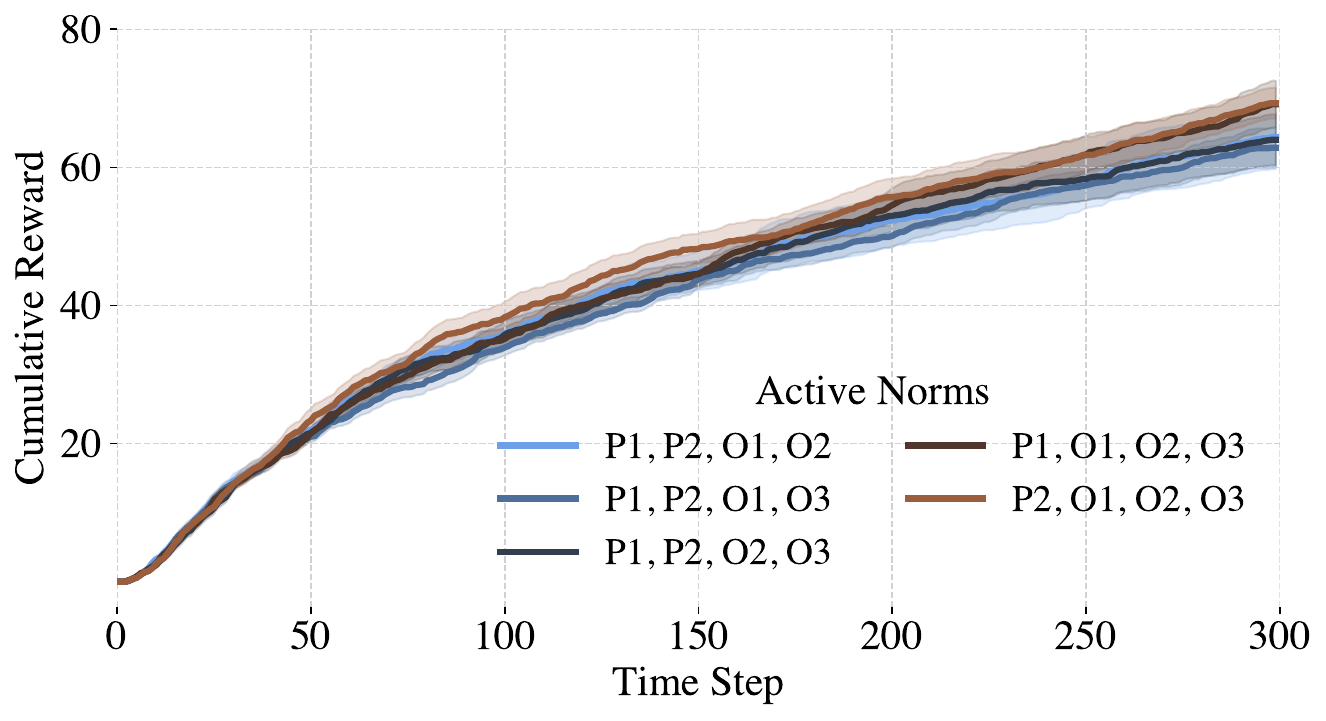}
    \subcaption{\textmd{Four norms combinations.}}
    \label{subfig:append_four_rule_desiccated}
  \end{subfigure}  
  \caption{\textmd{Cumulative reward per sets of practiced norms.}}
  \label{fig:append_learner_rewards}
\end{figure}

\begin{figure}[!h]
  \begin{subfigure}{0.45\linewidth}
    \centering
    \includegraphics[width=\linewidth]{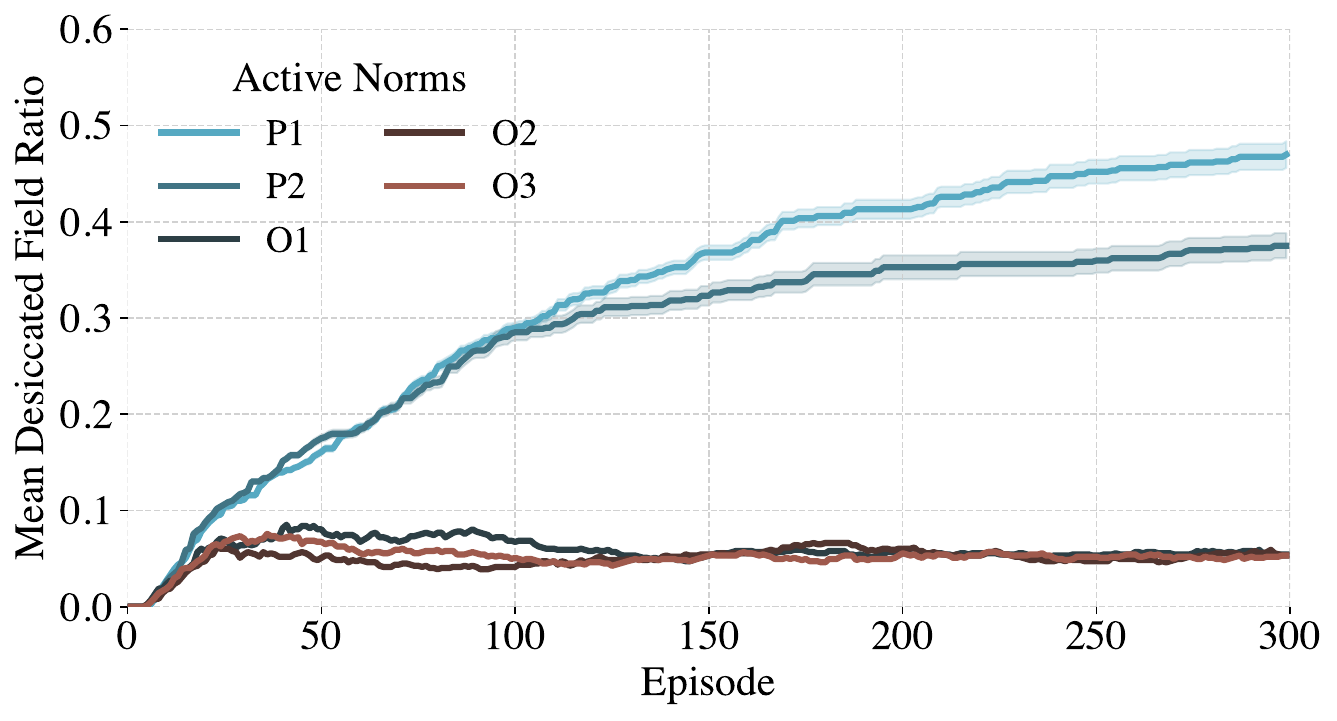}
    \subcaption{\textmd{One norm.}}
    \label{subfig:append_one_rule_desiccated}
  \end{subfigure}  
  \begin{subfigure}{0.45\linewidth}
    \centering
    \includegraphics[width=\linewidth]{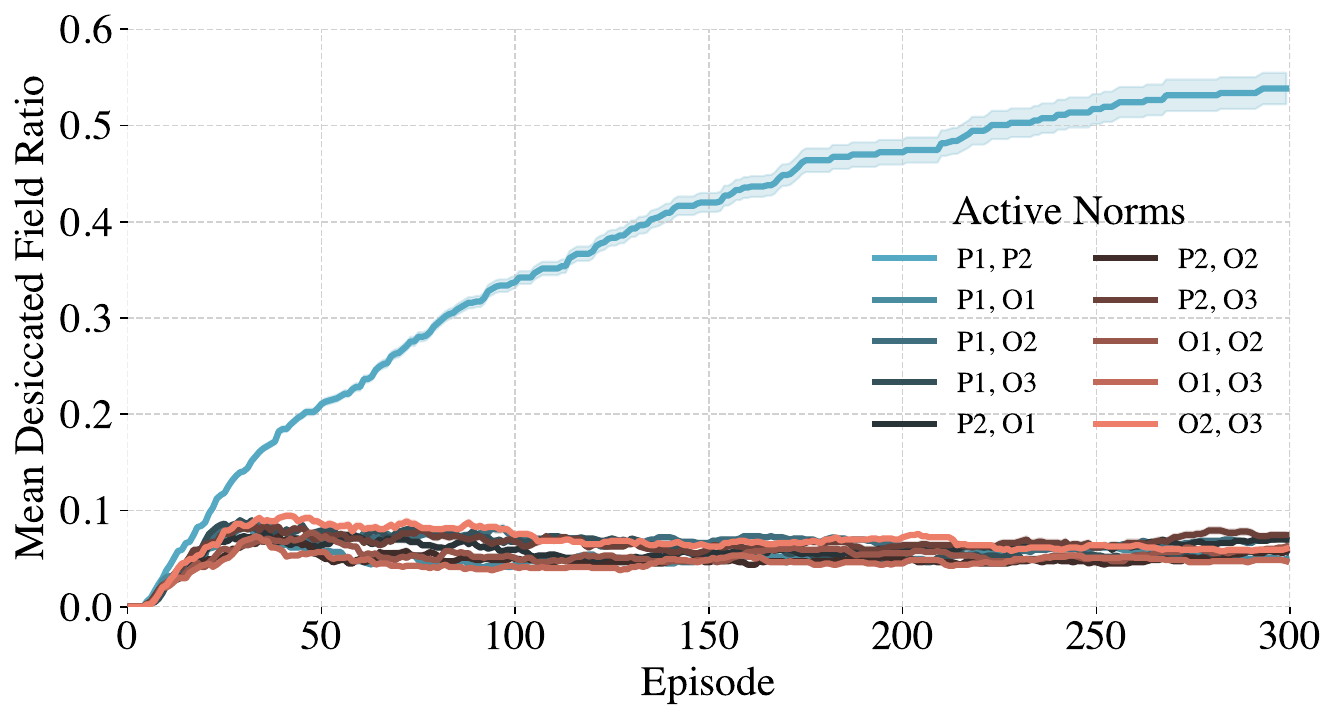}
    \subcaption{\textmd{Two norms combinations.}}
    \label{subfig:append_two_rule_desiccated}
  \end{subfigure}  
  \begin{subfigure}{0.45\linewidth}
    \centering
    \includegraphics[width=\linewidth]{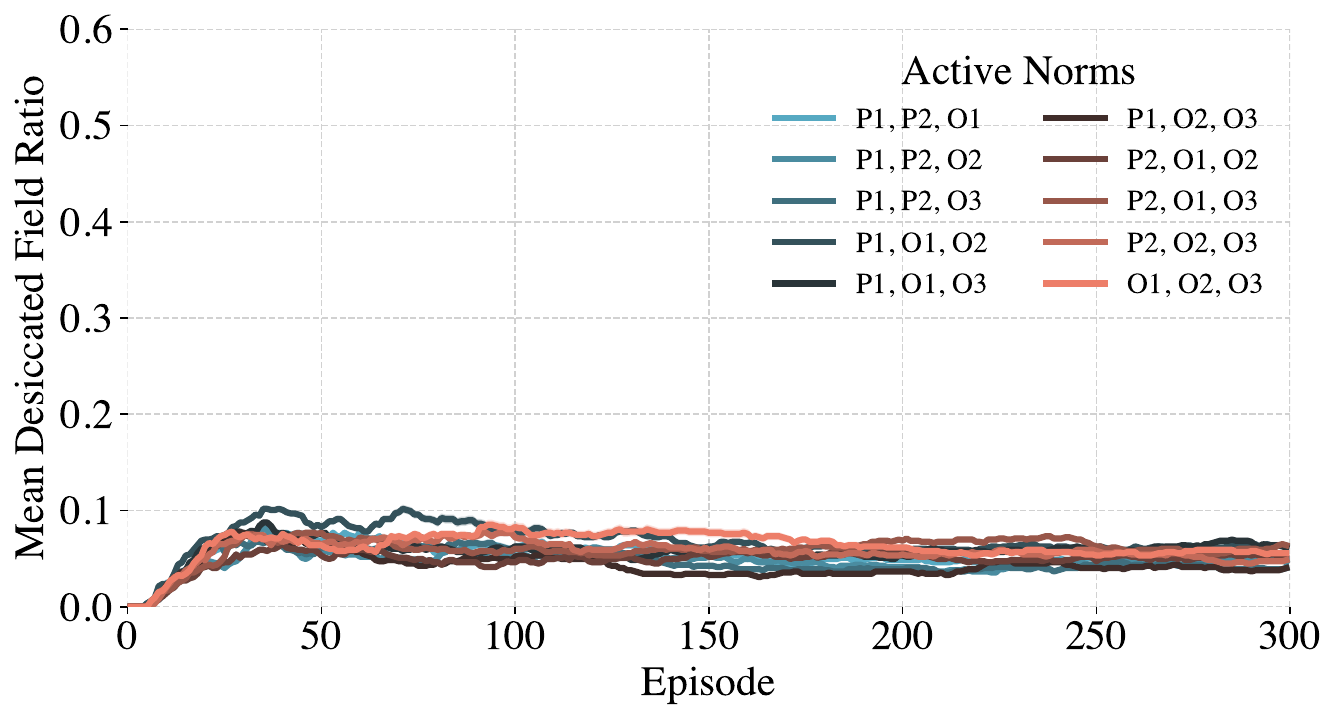}
    \subcaption{\textmd{Three norms combinations.}}
    \label{subfig:append_three_rule_desiccated}
  \end{subfigure}  
  \begin{subfigure}{0.45\linewidth}
    \centering
    \includegraphics[width=\linewidth]{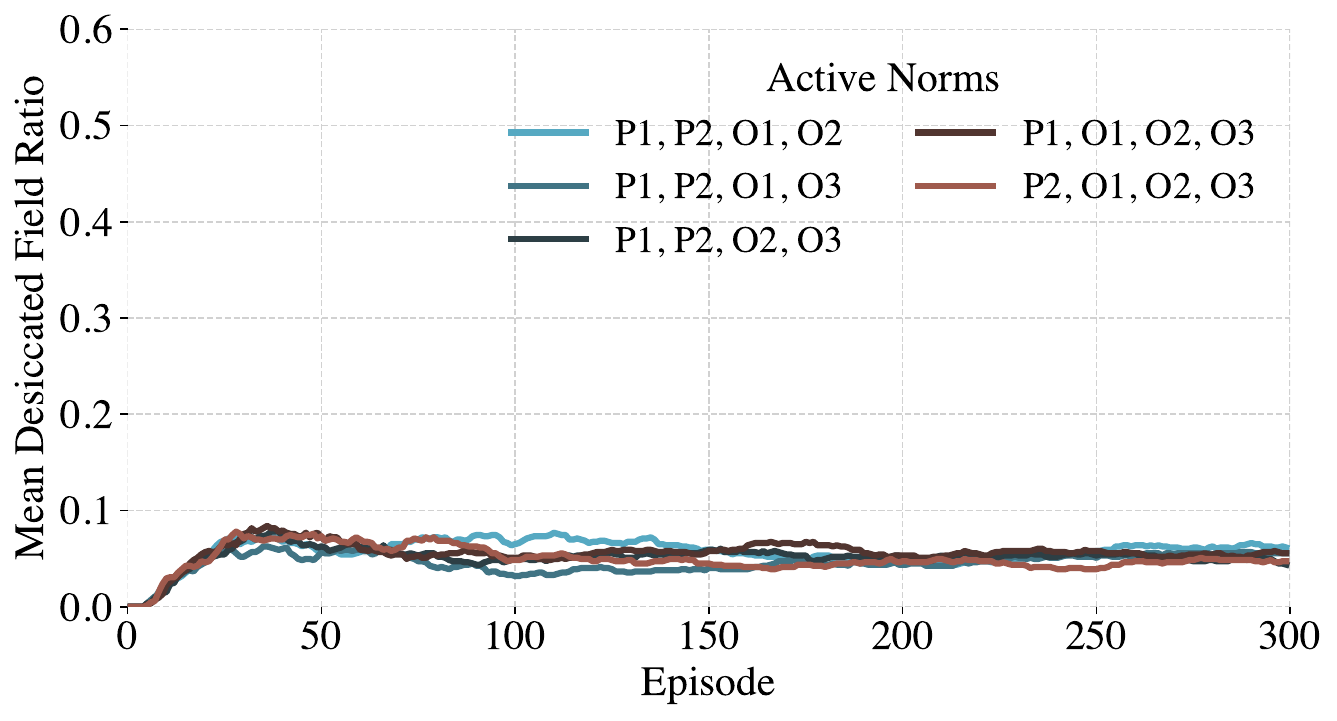}
    \subcaption{\textmd{Four norms combinations.}}
    \label{subfig:append_four_rule_desiccated}
  \end{subfigure}  
  \caption{\textmd{Desiccated field ratio per sets of practiced norms.}}
  \label{fig:append_mean_desiccated}
\end{figure}

\begin{figure}[!h]
  \begin{subfigure}{0.45\linewidth}
    \centering
    \includegraphics[width=\linewidth]{Graphics/too_few_apples_prohibition.pdf}
    \subcaption{\textmd{P1: Don't empty orchard.}}
    \label{subfig:append_do_not_empty}
  \end{subfigure}  
  \begin{subfigure}{0.45\linewidth}
    \centering
    \includegraphics[width=\linewidth]{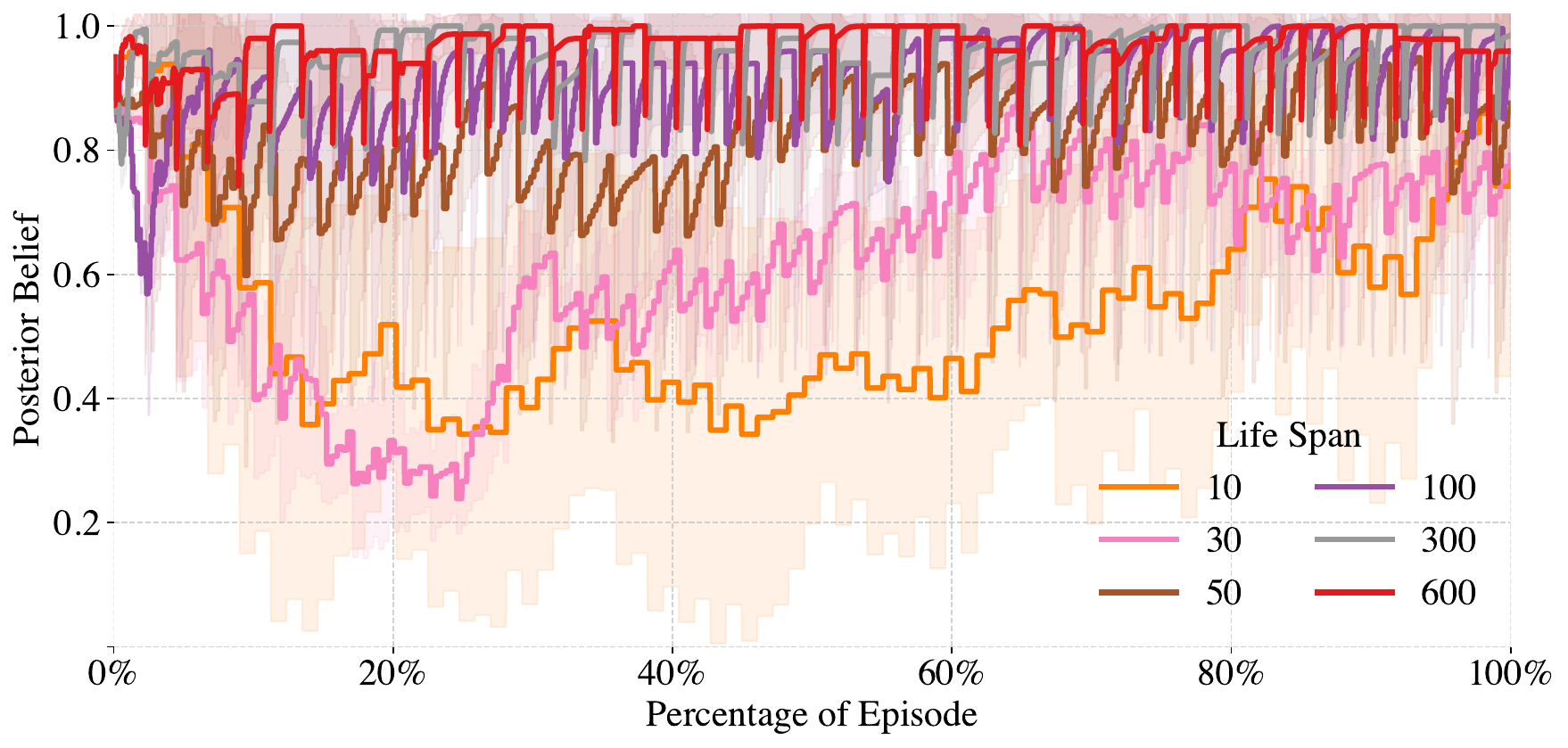}
    \subcaption{\textmd{P2: Don't steal.}}
    \label{subfig:append_do_not_steal}
  \end{subfigure}  
  \begin{subfigure}{0.45\linewidth}
    \centering
    \includegraphics[width=\linewidth]{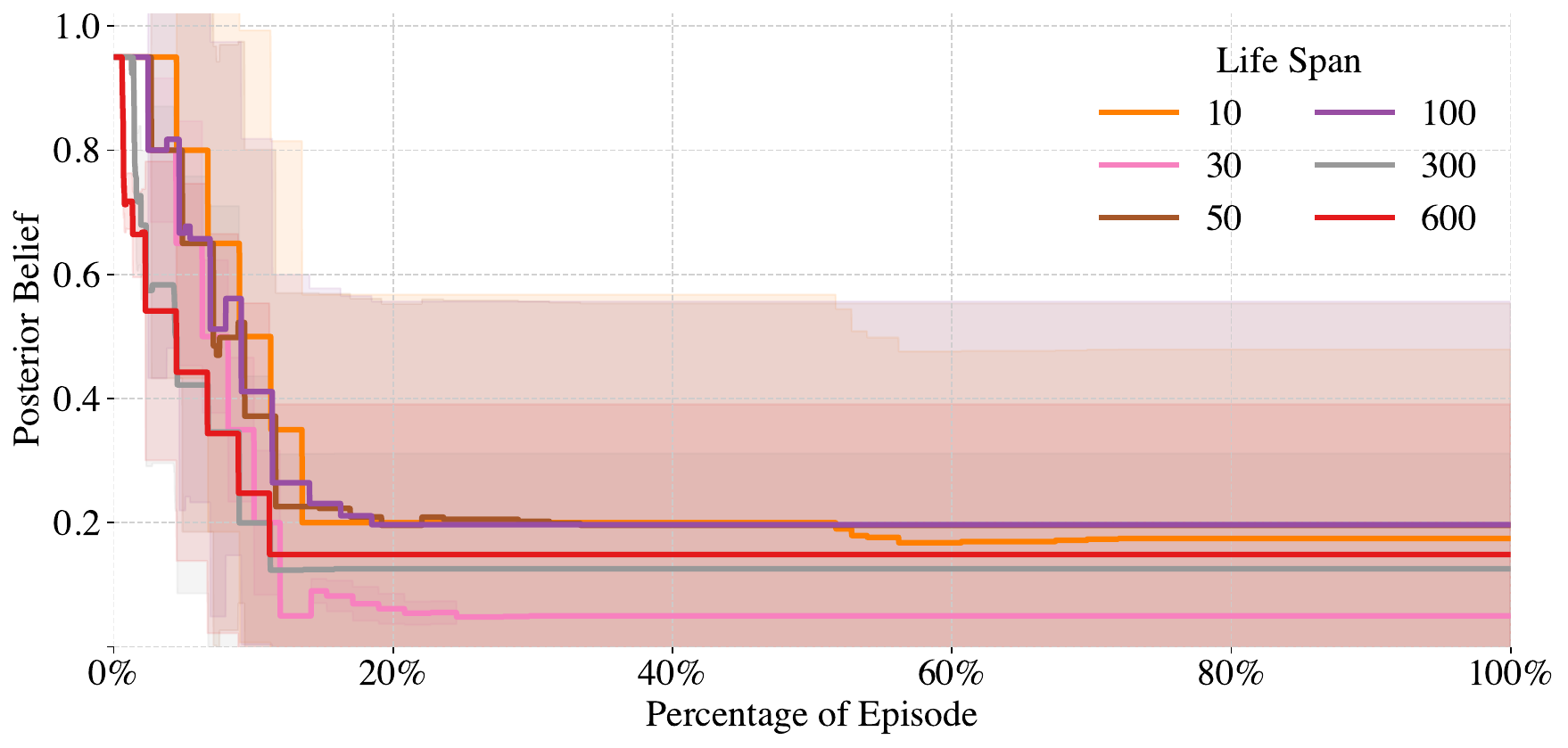}
    \subcaption{\textmd{O1: Farmer should pay.}}
    \label{subfig:append_farmer_obl}
  \end{subfigure}  
  \begin{subfigure}{0.45\linewidth}
    \centering
    \includegraphics[width=\linewidth]{Graphics/cleaner_clean_obligation.pdf}
    \subcaption{\textmd{O2: Cleaner should clean.}}
    \label{subfig:append_cleaner_obl}
  \end{subfigure}  
  \begin{subfigure}{0.45\linewidth}
    \centering
    \includegraphics[width=\linewidth]{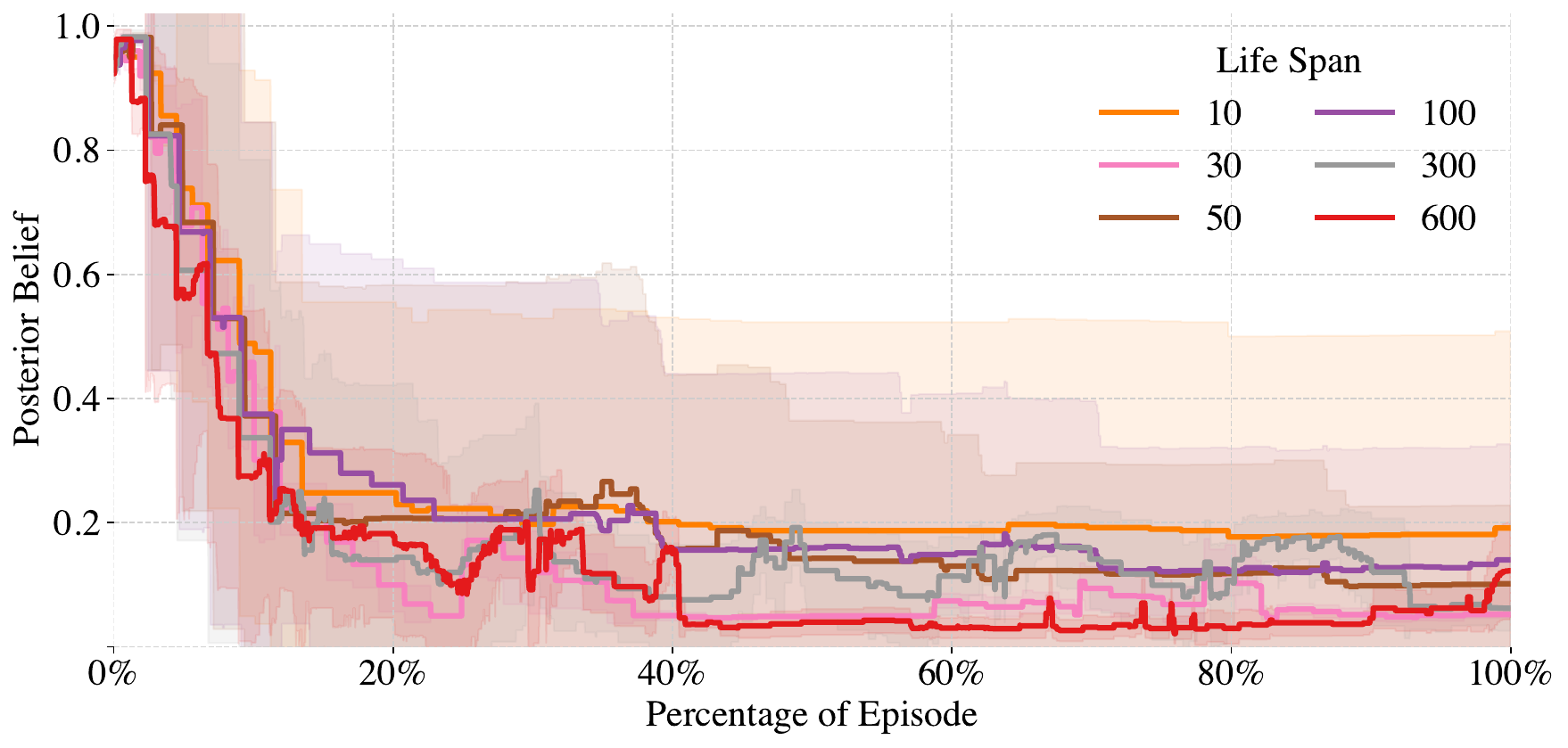}
    \subcaption{\textmd{O3: Egalitarian should clean.}}
    \label{subfig:append_egalitarian_obl}
  \end{subfigure}  
  \begin{subfigure}{0.45\linewidth}
    \centering
    \includegraphics[width=\linewidth]{Graphics/average_beliefs_across_life_spans.pdf}
    \subcaption{\textmd{Average}}
    \label{subfig:append_average_intergen}
  \end{subfigure}  
  \caption{Intergenerational transmission of norms. \textmd{Mean norm belief averaged across all six agents across generations. Figures \textbf{(a)}-\textbf{(e)} shows the single norms as described in Figure\ref{subfig:social_norms}. Figure \textbf{(f)} is the average of \textbf{(a)}-\textbf{(e)}.}}
  \label{fig:append_intergen}
\end{figure}

\clearpage
\begin{longtable}{rlc}
\caption{\textmd{Top 10 spontaneously emerging norms and percentage of their occurrence, averaged over 13 runs.}} \\
\label{tab:append_emerging_norms}
 & Norm & Occurs \\
1 &  If the river dirt fraction > 0.55 $\to$ Don't move. & 0.61\% \\
2 & If the river dirt fraction > 0.6 $\to$ Don't move. & 0.61\% \\
3 & If the river dirt fraction > 0.3 and you look like an egalitarian $\to$ Clean the river. & 0.61\% \\
4 & If you're looking towards East $\to$ Don't move. & 0.46\% \\
5 & If a cell holds an apple and the number of apples around < 1 $\to$ Don't move there. & 0.38\% \\
6 & If it's not your property $\to$ Don't move there. & 0.31\% \\
7 & If a cell holds an apple and it's not your property $\to$ Don't move there. & 0.31\% \\
8 & If a cell holds an apple and the number of apples around < 1 $\to$ Don't move there. & 0.31\% \\
9 & If you're looking towards West $\to$ Don't move. & 0.31\% \\
10 & If the river dirt fraction > 0.55 $\to$ Don't move. & 0.23\% \\
\end{longtable}

\end{document}